\documentclass[11pt]{article}
\pdfoutput=1

\usepackage[T1]{fontenc}
\usepackage{amsmath}
\usepackage{newtxtext}
\usepackage{newtxmath}
\usepackage[scaled=0.92,zerostyle=b]{newtxtt}
\usepackage{microtype}
\usepackage{array}
\usepackage{booktabs}
\usepackage{graphicx}
\usepackage[margin=1in]{geometry}
\usepackage[margin=1.5cm]{caption}
\usepackage{float}
\usepackage{xcolor}
\usepackage{url}
\usepackage[round]{natbib}
\usepackage[hidelinks]{hyperref}
\usepackage{tikz}
\usetikzlibrary{arrows.meta,positioning,calc}

\numberwithin{equation}{section}
\newcolumntype{L}[1]{>{\raggedright\arraybackslash}p{#1}}

\definecolor{fmaccent}{HTML}{2C5C7A}

\newif\ifreviewdraft
\reviewdraftfalse
\newcommand{\drafttodo}[1]{%
  \ifreviewdraft
  \par\medskip
  \noindent\fcolorbox{red!65!black}{yellow!12}{%
    \parbox{\dimexpr\linewidth-2\fboxsep-2\fboxrule\relax}{%
      \textbf{\color{red!65!black}DRAFT TODO.} #1}}
  \par\medskip
  \fi}

\begin{document}

\title{DraftFM: A Foundation Model for Day-Zero Drafting\\
in \emph{Magic: The Gathering}}
\author{Brian Ward\\
\addlinespace
  {\small Independent Researcher}\\
  {\small\texttt{brian.ward.92@gmail.com}}}
\date{August 10, 2026}
\maketitle

\begin{abstract}
Drafting a new \emph{Magic: The Gathering} expansion begins before any pick
from it has been observed: the complete card list is public, but the draft
logs that supervised pick models train on do not yet exist. We study this
day-zero regime directly. DraftFM is a discrete-choice policy that scores
exactly the cards available in the current pack, conditioned on the drafted
pool and the state of the draft. Every card enters as a frozen
775-dimensional function of its public card record, structured features and
a fixed text embedding, with no card identities, set identities, or usage
statistics anywhere in the model, so an unseen card is scored by the same
machinery as a familiar one. A 1.6-million-parameter network fitted on 149
million human picks from 29 expansions predicts held-out picks in three
expansions withheld in their entirety, reaching 50.8\%, 60.4\%, and 56.7\%
top-1 agreement, where uniform chance at the opening pick is about 7\%.
Refitted on all 32 observed expansions, the same architecture produced a
card ranking for the then-unreleased set \emph{The Hobbit}, sealed with its
complete cryptographic provenance and published roughly 36 hours before the
set became draftable on MTG Arena. The sealed ranking agrees with six
independent expert reviewers roughly as much as those reviewers agree with
one another. Evaluation against realized outcomes is committed to a
follow-on note, whatever it shows.
\end{abstract}


\newpage

\tableofcontents

\newpage

\section{Introduction}
\label{sec:introduction}

Drafting in Magic: The Gathering is a sequential game of imperfect information.
Eight players open packs of cards and pass them around the table, each privately
assembling a deck one pick at a time. Every few months a new set of several
hundred cards is released, and drafters must learn it from scratch. For a
data-driven assistant this creates a day-one problem: supervised pick models
train on public draft logs, and those logs appear on a delay of roughly two weeks
after a set's release \citep{brooks2024statistical}. The players who most want
help with a new set are the ones no trained model can yet serve.

On release day, the state of the art is human judgment. Draftsim publishes
hand-tuned ratings written by human experts \citep{draftsimratings}, and drafters
consult them because nothing learned is available. The learned literature is a
within-set literature. It begins with \citet{ward2021}, who compare drafting
agents on Core Set 2019: a random agent takes the human's card 22\% of the time,
a bot driven by expert-tuned ratings 44.5\%, and a neural network trained on
human drafts 48.7\%. Its successors define the deployed state of the art. The
statistical-drafting models of \citet{brooks2024statistical}, multilayer
perceptrons trained per set, reach roughly 70\% pick agreement, and the puder
transformer of \citet{czerner2025puder} reports 71\%. All of these models encode
cards as vocabulary indices. Each learns from picks made in the one set it
serves, so none can score an unseen set, and none can exist until the two-week
data lag has passed.

Large language models look like a way around the lag, because they can read a new
card's rules text the day it is published. \citet{bertram2025} tests this
directly. Zero-shot GPT-4o agrees with human picks 43\% of the time on Kamigawa:
Neon Dynasty, below the expert-tuned ratings above, and supplying each card's
full rules text in the prompt lowers its accuracy further. Open 7 to 8B models fail
to pick legal cards at all until they are tuned on a million picks from the
target set, after which they still trail small supervised multilayer perceptrons
at far higher inference cost. Even the 43\% figure carries a caveat: the set was
released years before the model's training cutoff, so post-release discussion of
its strategy cannot be ruled out of the pretraining corpus, a leak a frozen
feature-only encoder cannot have. The design lesson we take is to use a frozen
text embedding as a feature, not a language model as the policy.

This paper goes further to ask how far pure transfer can go, and it isolates the day-zero
question by construction. DraftFM is a single pick model trained across public
17Lands draft data. To DraftFM, a card is nothing but a frozen vector of public
information: structured features parsed from its Scryfall record, plus a sentence
embedding of its rules text. The model has no set-identity parameters and no
per-card weights. Whatever it knows about a set it must infer from the cards in
the pack and the drafter's accumulating pool, and that inference transfers to a
set released tomorrow, which a memorized card vocabulary cannot.

The nearest prior work is \citet{bertram2024a}, who established the problem of
drafting a set the model never trained on and showed that multi-set pretraining,
not the representation alone, is what buys transfer. Pretrained on thirteen sets
and evaluated on held-out The Brothers' War, their best representation reaches
55.4\% pick agreement. That representation includes a block of post-release
usage statistics computed from human play, so 55.4\% is an anchor for
generalizing across cards rather than for day-one deployment. Their
day-one-feasible representation, features only, reaches 33.6\% to 35.6\% on unseen
sets under single-set training, and no multi-set, features-only number was
published. Filling that gap is this paper's first contribution. We provide
a feature-only policy trained on 29 expansions and evaluated zero-shot on three
held-out and pre-selected expansions. Our feature set includes no usage statistics
from the target set, providing a day-zero draft advisor. To compare to existing
literature, we chose The Brothers' War as in \citet{bertram2024a}, as well as
the recent Marvel Super Heroes set, and Magic: Foundations, which was released
roughly at the midpoint of BRO and MSH's release dates and carries some of the
core themes and mechanics of MTG.

The second contribution is a forecast that could not be fitted after the fact. We
sealed and publicly published a complete pack-1-pick-1 ranking of a new set
before it was playable on MTG Arena, with the digests of its inputs, its model,
and its outputs, and we commit in advance to the evaluation that will score it
once the draft logs appear. The third is a comparison against the only forecasts
that exist at that moment: the pre-release set reviews published by Limited
content creators, placed on a common ladder and compared to each other and to the
model on the same footing.

These numbers should not be read across regimes. A within-set model with access
to its target set's own picks is solving an easier problem than a model that has
never seen the set, and the gap between the two is the subject of this paper
rather than a defect in either. Our own in-distribution validation agreement is
68.6\%, squarely inside the 66--71\% band the within-set literature reports, and
that figure is not comparable to our zero-shot results. The 55.4\% zero-shot
figure above is likewise not a target we are trying to beat, because it uses
human-usage statistics that do not exist on day zero. The development numbers we
report on three whole expansions answer a different question: what a policy can
do on a set for which nothing has yet been observed beyond its contents?

The rest of the paper is organized as follows.
Section~\ref{sec:data-modeling} describes the two data sources, the card
representation, and the model. Section~\ref{sec:development-evaluation} gives the
evaluation design, compares three model widths on three whole held-out
expansions, and analyzes how predictive behavior varies across the draft.
Section~\ref{sec:hob-forecast} reports the sealed pre-release forecast, its
publication record, and the comparison with content creators.
Section~\ref{sec:conclusion} concludes. Appendices cover the public artifact and
the complete development results.

\section{Data and Modeling Setup}
\label{sec:data-modeling}

\subsection{Data Sources}
\label{sec:data-sources}

DraftFM uses two public data sources. 17Lands supplies observed human choices
and the state of the draft in which each choice was made. Scryfall supplies the
printed characteristics of the cards. Table~\ref{tab:data-source-roles} gives
this division of labor.

\begin{table}[tb]
\centering
\caption{The two data sources used by DraftFM.}
\label{tab:data-source-roles}
\begin{tabular}{L{0.15\textwidth}L{0.20\textwidth}L{0.55\textwidth}}
\toprule
Source & Unit & Information used \\
\midrule
17Lands & One observed draft pick & The cards offered in the pack, the cards
already drafted, the selected card, the pick's location in the draft, and
available drafter and event information \\
\addlinespace
Scryfall & One card printing & Card identity, mana cost and value, colors,
rarity, type line, rules text, numerical characteristics, layout and card-face
structure, and set metadata \\
\bottomrule
\end{tabular}
\end{table}

17Lands collects MTG Arena logs from players who opt in through its tracker and
publishes anonymized bulk datasets for research and community analysis.\footnote{See
\url{https://www.17lands.com/public_datasets} for the datasets
published by 17Lands.} We use its \texttt{draft\_data} files. Each row records
one card selected by a tracked drafter. It includes the chosen card, the pack
and pick numbers, draft and event identifiers, rank, and available outcome
fields. It also includes two groups of card-count columns. A
\texttt{pack\_card\_<NAME>} column says how many copies of that card were in the
current pack. A \texttt{pool\_<NAME>} column says how many copies the player had
already drafted. Together these columns record the choice set, the pool at the
time of the choice, and the card taken. Unless otherwise noted, 17Lands
publishes these files under the Creative Commons Attribution 4.0 International
license. We attribute them as ``Data from 17Lands.com (CC BY 4.0).''

For each set and draft format, we download the gzip-compressed CSV published in
the 17Lands public Amazon S3 bucket. Files are fetched once, following the
17Lands usage guidelines for bulk downloads, and kept together with their source
metadata and a content hash, which lets us identify the exact input used in an
experiment. The acquisition step records source metadata and a
content hash for every downloaded file, and a later step freezes those hashes
into a manifest that subsequent stages verify against.

Our working collection contains 60 set-format files from 32 draft sets. It
begins with \emph{Strixhaven: School of Mages} (STX in April 2021) and extends
through \emph{Marvel Super Heroes} (MSH in June 2026). 17Lands publishes
Premier and Traditional Draft as separate files, so most sets contribute two.
Four do not: AFR, MID, MSH, and VOW have no Traditional Draft file in the
collection. The 60 files are therefore 32 Premier Draft files, one per set, and
28 Traditional Draft files. Together they contain 169,932,378 observed
picks. After combining formats within each expansion, the mean is 5.31 million
picks per set, the sample standard deviation is 2.48 million, and the median is
4.84 million. Individual sets contribute between 1.07 and 10.60 million picks.
BRO, FDN, and MSH are withheld in their entirety as whole-set development
environments for zero-shot model comparison. They contain 12,571,237 picks
that are never used for fitting or within-training validation. The remaining
29 sets contain 157,361,141 picks. A deterministic split by
draft identifier assigns 149,483,436 of them to fitting and 7,877,705 to internal
validation. The experimental-design section gives the complete protocol.
HOB is not part of the observed-pick collection. At forecast time, HOB
contributes only its public Scryfall records and the predictions sealed from
them.

The raw 17Lands files are wide tables, with a separate pack and pool column for
every card in a file's vocabulary. Their metadata schema changed during the
observation period. Early files use match-based skill-bucket names, some
intermediate files call the rank field \texttt{user\_rank}, and modern files
use game-based bucket names and \texttt{rank}. We map equivalent fields into
one canonical schema, leave their recorded values unchanged, and insert typed
nulls when an optional field did not yet exist. We then store the tables as
compressed Parquet. This makes the same information readable in the same way
across all 32 sets.

We also save the ordered card vocabulary defined by the
\texttt{pack\_card\_} columns and encode the card named in \texttt{pick} as its
zero-based index in that vocabulary. This \texttt{pick\_index} identifies the
card selected by the drafter. Draft position remains in \texttt{pack\_number}
and \texttt{pick\_number}. These operations standardize storage. They neither
construct model features nor alter the observed pack, pool, or choice.

Scryfall is a community-maintained MTG card database with a public
API.\footnote{See \url{https://scryfall.com/docs/api/bulk-data} for Scryfall's
bulk-data documentation.} We use its
\texttt{all\_cards} bulk object for card identities and publicly observable card
attributes, not for human choices or draft outcomes.
The acquisition step queries the bulk-data endpoint,
selects that object, follows the download URL supplied by the API, and retains
a timestamped snapshot. The present endpoint supplies a gzip-compressed
JSON Lines file with one JSON object per printing. The downloader also accepts
the earlier JSON-array representation. Retaining the complete dated snapshot
allows every run to use the same card information after Scryfall itself is
updated. The snapshot used in this study was retrieved on August 8,
2026 at 13:35:59 EDT and stored as
\texttt{all\_cards\_20260808133559.json}. Its SHA-256 digest is
\texttt{4a60c20e800c5d55d2459e8edc19355bee794873fc84d653999a145fe9ac130f}.

We keep the English-language records and write the three Parquet tables
\path{cards.parquet}, \path{card_faces.parquet}, and \path{sets.parquet}.
Table~\ref{tab:scryfall-fields} lists the fields used
either to resolve a 17Lands card name to one reproducible Scryfall printing or
to construct its semantic representation. Identity and release fields control
that join. They do not enter the network as numerical card features. Image and
price fields are retained for other applications but are not DraftFM inputs.
Section~\ref{sec:card-representation} describes the transformations from these
source fields to model features.

\begin{table}[tb]
\centering
\caption{Scryfall fields used for identity resolution and card features.}
\label{tab:scryfall-fields}
\begin{tabular}{L{0.18\textwidth}L{0.20\textwidth}L{0.52\textwidth}}
\toprule
Record & Field group & Fields \\
\midrule
Card printing & Identity resolution & Scryfall identifier, name, set code,
digital status \\
\addlinespace
Card printing & Card features & Mana cost, mana value, colors, color identity \\
\addlinespace
Card printing & Card features & Rarity, type line, rules text, keywords, layout \\
\addlinespace
Card printing & Card features & Power, toughness, loyalty \\
\addlinespace
Card face & Card features & Face index, name, mana cost, type line, rules
text, colors, power, toughness, loyalty \\
\addlinespace
Set & Printing selection & Set code, release date \\
\bottomrule
\end{tabular}
\end{table}

17Lands identifies cards by name, whereas Scryfall may contain several records
for the same card because it has been reprinted or issued with different art.
We normalize the 17Lands name and choose the corresponding Scryfall printing,
preferring the printing from the same expansion. Full, front-face, and
back-face names are recognized for multi-face cards. If a name cannot be
matched, processing stops so that the missing card can be corrected. The
result gives every card in a 17Lands pack or pool one Scryfall description.

\subsection{DraftFM}
\label{sec:draftfm}

\subsubsection{Behavioral Prediction, Transfer, and Prior Work}
\label{sec:behavioral-transfer}

A draft pick is a choice among the cards remaining in the pack. The choice is
conditioned on the cards already drafted, the
location of the pick, the event format, and the drafter's observed experience.
DraftFM estimates the probability that a human drafter selects each available
card under those conditions. It is therefore a flexible discrete-choice model.
The neural network replaces the linear utility function of a conventional
conditional logit model, while the output remains a softmax over exactly the
cards that were available \citep{mcfadden1974}.

This target is behavioral. Agreement with a held-out pick tells us whether the
model anticipated the action of a human drafter. It does not, by itself, show
that the predicted card maximizes win rate or that the human pick was correct.
DraftFM is trained on picks from players at different observed skill and
experience levels and includes those quantities as conditioning variables. For
deployed recommendations, we set these variables to describe a high-skill
drafter. The resulting policy estimates choices associated with stronger
drafters, but it remains a model of observed behavior rather than an ``optimal
draft model.''

\citet{ward2021} established the modern behavioral benchmark for MTG drafting.
The task is to predict the card a human selected from the available pack and
current pool. They collected 107,949 simulated human
drafts of M19, trained and tested within that single expansion, and compared
random, rarity-based, expert-tuned, Bayesian, and neural agents. Their neural
agent achieved 48.67\% top-1 agreement, the strongest result in the comparison.
It encoded the pool as a 265-dimensional vector of M19 card counts, scored all
265 card identities, and only then masked the scores to the current pack. Its
per-pick analysis also established an important measurement fact. Raw agreement
is highest near the beginning and end of each pack and lower in the middle.
Pack size and consensus therefore change together over a draft, so an aggregate
accuracy alone cannot establish that an agent is using context well.

DraftFM keeps their behavioral target but replaces the two fixed coordinate
systems that prevent transfer. Those systems are the identity-indexed pool and
identity-indexed output. Every pooled and offered card is instead represented by shared features
and processed by the same learned functions. The output softmax therefore has
one entry per available card, not one entry per card identity in a particular
set. One fitted policy can evaluate cards that did not exist during training.

\citet{bertram2021} supplied the next essential step by making
pool context part of the learned representation. Their contextual preference
ranking model learns that the observed pick is a better addition to the current
pool than each rejected card. That study still used one coordinate
per M19 card. Their later work replaces those identities with generalized
representations assembled from structured card attributes, text, images, and
human-usage statistics \citep{bertram2024a,bertram2024b}. They show decisively that semantic
representations are required to score unseen cards meaningfully and that
training across sets improves transfer. Their multi-set model, trained on 75
million picks, reached 55.44\% top-1 agreement on the single withheld expansion
BRO.

That result is the closest predecessor to DraftFM, but it does not isolate the
day-zero multi-set question. The reported multi-set model combines card
features and images with sixteen statistics derived from human use, including
pick and game outcomes. Those statistics do not exist before a set is played.
The paper tests a feature-only representation in a model trained solely on NEO,
but does not report a feature-only model trained across multiple sets and
evaluated zero-shot on a whole expansion. DraftFM isolates exactly that experiment.
It trains one policy without target-set picks or performance statistics and
evaluates it on three expansions withheld in their entirety.

Recent systems reinforce the distinction between specialization and transfer.
The \emph{statistical-drafting} models are refreshed from a new set's
17Lands data, while the \texttt{puder} transformer learns tokens for a fixed
historical card vocabulary.\footnote{See
\url{https://github.com/danieljbrooks/statistical-drafting} and
\url{https://nicze.de/philipp/articles/puder/}.} Both can achieve strong
in-distribution agreement, but neither can assign a learned token to a genuinely
unseen card on day zero. UrzaGPT instead expresses the pack and pool as language
and fine-tunes a 7--8 billion parameter language model. It reaches 66.2\%
agreement after training on one million NEO picks, while its author explicitly
leaves cross-expansion generalization open \citep{bertram2025}. This
demonstrates that
language is a viable card interface. It does not test a learned policy on a new
expansion. Cardsformer reaches the analogous conclusion in Hearthstone, using
language representations so a game-playing policy can act on unseen
cards \citep{xia2023}.

Draft outcome models address a complementary question. \citet{rigaux2026} use
set-contextualized card embeddings and the full pick sequence to predict the
eventual deck's win rate, rather than the drafter's next choice. Their results show that a draft sequence
contains information about downstream performance. DraftFM models the choices
that create that sequence. Among the published systems above, it is the first
to combine multi-expansion behavioral training, semantic card inputs, explicit
pool context, and multiple whole-expansion zero-shot evaluations without
target-set usage statistics.

The \emph{FM} in DraftFM stands for \emph{foundation model}. A foundation model
is trained broadly and reused across downstream applications
\citep{bommasani2021}. DraftFM is a foundation model for
MTG drafting because one policy is trained across many draft environments and
reused when the available card set changes. Its foundation role is operational. The same
fitted model supplies P1P1 card ratings, context-dependent pick probabilities,
and controlled analyses of how predicted behavior changes with the state of a
draft. Target-set fitting is a separate and easier regime because it adds live
information that is unavailable at release. We evaluate it separately rather
than weakening the day-zero test.

Here, \emph{day zero} means that the complete public card list is available but
no picks from the target set are used. The prediction is therefore independent
of the target set's live pick counts, win rates, usage statistics, community
ratings, and fitted card-identity parameters. This is the distinction that
permits the sealed HOB forecast.

``Unseen'' describes the set, not every card in it. A held-out expansion's packs
include some ordinary reprints, mostly staples and basic lands, that also appear
in the fitting corpus under other sets. This confers no advantage under the
contract above, because a reprint has no memorized identity to exploit, only the
same frozen feature vector any novel card with identical features would receive.
It does mean that set-level and card-level novelty are not the same claim.

Context also separates DraftFM from a static set review. At P1P1 the drafted
pool is empty, so querying all cards under the same conditions produces a
general card ranking. At later picks the pool encoder can change a card's score
in response to colors, repeated effects, curve requirements, and other learned
card interactions. The accumulating pool therefore gives the model a
progressively richer input. Later analyses test whether the model extracts
predictive value from that information beyond the mechanical increase in raw
agreement caused by shrinking packs.

Because the target is observational, the behavioral analyses in this paper are
descriptive. For example, holding a pack fixed and varying the pool or skill
condition can reveal an association learned by the model. It does not identify
the causal effect of possessing a card or becoming a stronger player. We use
held-out prediction and controlled model queries to distinguish what the model
has learned without giving those comparisons a causal interpretation.

\subsubsection{Card Representation, Neural Architecture, and Fitting}
\label{sec:card-representation}

Each 17Lands row becomes one supervised choice problem. The cards with positive
\texttt{pack\_card\_} counts form the available alternatives. The
\texttt{pool\_} counts describe the state before the choice. The recorded
pick is the target. Cards are joined by name to Scryfall and then replaced by
frozen feature vectors. Consequently, the training target comes only from
17Lands, while the description of every candidate and pooled card comes only
from Scryfall.

Every card is represented by 775 numbers. Of these, 391 are structured features
and 384 form a text embedding. Table~\ref{tab:card-feature-blocks} gives the
structured blocks. The design favors ordinary card properties that have the
same meaning across expansions. It contains no card-identity, set-identity,
artist, price, image, or post-release performance feature.

\begin{table}[tb]
\centering
\caption{The structured component of the DraftFM card representation.}
\label{tab:card-feature-blocks}
\begin{tabular}{L{0.28\textwidth}rL{0.50\textwidth}}
\toprule
Feature block & Dimensions & Contents \\
\midrule
Mana value & 10 & Scaled value and buckets from zero through eight or more \\
Mana cost & 10 & Colored and colorless pips, generic mana, and indicators for
$X$, hybrid, and Phyrexian costs \\
Color & 8 & Five colors, number of colors, colorless, and multicolored \\
Color identity & 5 & WUBRG color-identity indicators \\
Supertypes and types & 12 & Legendary, snow, basic, and nine card types \\
Creature subtypes & 129 & Up to 128 reserved vocabulary slots and an unmatched count \\
Numerical characteristics & 9 & Scaled power, toughness, and loyalty with
missing and variable-value indicators \\
Rarity & 5 & Common, uncommon, rare, mythic, and other \\
Keyword abilities & 167 & Up to 166 reserved vocabulary slots and an unmatched count \\
Layout & 10 & Layout class and double-faced-card indicators \\
Rules-text shape & 26 & Text length, line count, and fixed indicators for common
functional patterns \\
\midrule
Total & 391 & \\
\bottomrule
\end{tabular}
\end{table}

Two data-dependent vocabularies occur in this representation. The architecture
reserves 128 subtype slots and 166 keyword slots, but a fitted manifest may
populate fewer than the reserved number. The populated entries are selected
from training-set cards and then frozen. Any unfilled reserved columns remain
zero. In the final all-data manifest described in
Section~\ref{sec:final-refit-protocol}, the subtype vocabulary is populated to
capacity, filling 128 of its 128 reserved slots, while the keyword vocabulary
populates 145 of its 166 reserved slots and leaves the remaining 21 keyword
columns permanently zero. The reserved counts in
Table~\ref{tab:card-feature-blocks} are therefore the fixed width of the
representation, not a count of distinct subtypes or keywords the model has seen.
Unrecognized subtypes and keywords in a held-out set contribute to
unmatched counts rather than creating new fitted columns. Numeric values are
scaled and clipped, and an unavailable or unparseable characteristic receives
an explicit missing indicator rather than silently becoming a meaningful zero.
For multi-face cards, the numerical blocks describe the front face and explicit
features record the presence and broad type of the back face.

The remaining 384 values are an $L_2$-normalized embedding of the card's type
line and rules text from the frozen \texttt{BAAI/bge-small-en-v1.5} sentence
encoder, which is described by \citet{xiao2023}. A card's own name is replaced before
embedding so that a known proper noun cannot act as a hidden card identifier.
Reminder text is retained because it may be the only definition available for
a new mechanic, and the back face is appended when one exists. The sentence
encoder is not fitted on draft picks. Its output is computed once per card and
then held fixed during DraftFM training.

These exclusions define the transfer experiment as much as the included
features do. A per-card parameter can memorize that a particular rare is
strong. A target-set pick-rate feature can summarize thousands of people
discovering the same fact. DraftFM receives neither. It must learn reusable
relationships between observable card descriptions and human choices. New-set
prediction is possible because a Scryfall record produces the same 775 inputs
whether or not that card appeared in the training corpus.

Let $A$ denote the cards actually available in the current pack and let
$K=|A|$. The candidate input is a matrix $X_A\in\mathbb{R}^{K\times775}$, not
one flattened vector with a fixed number of columns. A 14-card P1P1 pack
therefore contains $14\times775=10{,}850$ scalar values arranged as 14
instances of the same 775-feature representation. After one card is removed,
a 13-card P1P2 pack contains $13\times775=10{,}075$ such values. There is no
requirement that every choice supply at least 10,850 values.

The same card encoder $f_{\mathrm{card}}$ is applied row by row, with shared
weights for every card and every pack position. It maps each raw vector through
a $775$--$512$--$d$ multilayer perceptron to a learned representation
$e_i\in\mathbb{R}^d$. Thus the encoded pack has shape $K\times d$. This is a
supervised representation learned for pick prediction, not an autoencoder.
The network is never asked to reconstruct the 775 inputs. In the implementation
the complete card list for a set is encoded once per homogeneous training
batch, after which the pack simply indexes the appropriate rows of that shared
table.

Rectangular tensors are used only to batch packs efficiently. Each row has 16
storage positions, enough for the largest supported pack, and positions beyond
the $K$ real candidates carry a \texttt{PAD} marker. A padded position is not a
null card and does not contribute a learned feature vector. Its output score is
masked to $-\infty$ before normalization and its choice probability is exactly
zero. The padding pattern is not passed to the card encoder or candidate scorer
as a predictive feature. Pack and pick number enter separately through the
context pathway described below.

The network separately summarizes the accumulated pool $P$ and combines that
summary with the state of the draft. For a real candidate $i\in A$, the
computation can be written as
\begin{align}
e_i &= f_{\mathrm{card}}(x_i), \\
p &= f_{\mathrm{pool}}\left(\{(e_j,q_j):j\in P\}\right), \\
h &= f_{\mathrm{context}}(\text{position},\text{format},\text{skill},
\text{shape}), \\
u_i &= f_{\mathrm{score}}([e_i;p;e_i\odot p;h]), \\
\Pr(Y=i\mid A,P,h) &= \frac{\exp(u_i)}{\sum_{k\in A}\exp(u_k)}.
\end{align}

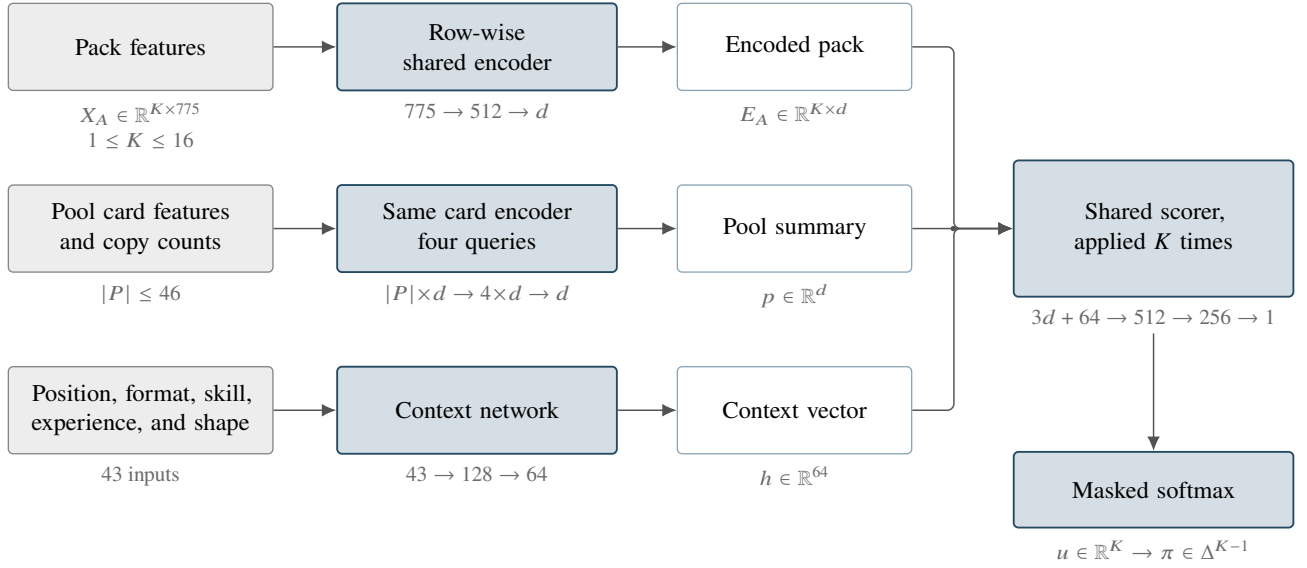
\begin{figure}[tb]
\centering
\begin{tikzpicture}[
  x=1cm, y=1cm,
  font=\footnotesize,
  note/.style={font=\scriptsize, text=black!58, align=center,
    inner sep=0pt, node distance=1.5mm},
  boxbase/.style={draw, rounded corners=1.8pt, align=center,
    inner xsep=2.5mm, inner ysep=1.6mm, minimum height=11.5mm,
    line width=0.5pt},
  datum/.style={boxbase, fill=black!7, draw=black!48, text width=30mm},
  learned/.style={boxbase, fill=fmaccent!20, draw=fmaccent!80!black,
    line width=0.7pt, text width=32mm},
  tensor/.style={boxbase, fill=white, draw=fmaccent!55, text width=26mm},
  flow/.style={-{Latex[length=1.9mm,width=1.7mm]}, line width=0.55pt,
    draw=black!68}
]
\node[datum]   (candidate)      at (1.75, 0)      {Pack features};
\node[learned] (cardencoder)    at (6.20, 0)      {Row-wise\\shared encoder};
\node[tensor]  (embedding)      at (10.40, 0)     {Encoded pack};

\node[datum]   (pool)           at (1.75,-2.40)   {Pool card features\\and copy counts};
\node[learned] (pooler)         at (6.20,-2.40)   {Same card encoder\\four queries};
\node[tensor]  (poolsummary)    at (10.40,-2.40)  {Pool summary};

\node[datum]   (conditions)     at (1.75,-4.80)   {Position, format, skill,\\experience, and shape};
\node[learned] (context)        at (6.20,-4.80)   {Context network};
\node[tensor]  (contextsummary) at (10.40,-4.80)  {Context vector};

\node[learned, text width=32mm, minimum height=18mm]
  (scorer)  at (15.15,-2.40) {Shared scorer,\\applied $K$ times};
\node[learned, text width=32mm, minimum height=10mm]
  (softmax) at (15.15,-5.85) {Masked softmax};

\node[note, below=of candidate.south]      {$X_A\in\mathbb{R}^{K\times775}$\\$1\leq K\leq16$};
\node[note, below=of cardencoder.south]    {$775\rightarrow512\rightarrow d$};
\node[note, below=of embedding.south]      {$E_A\in\mathbb{R}^{K\times d}$};
\node[note, below=of pool.south]           {$|P|\leq46$};
\node[note, below=of pooler.south]         {$|P|\!\times\!d\rightarrow4\!\times\!d\rightarrow d$};
\node[note, below=of poolsummary.south]    {$p\in\mathbb{R}^d$};
\node[note, below=of conditions.south]     {$43$ inputs};
\node[note, below=of context.south]        {$43\rightarrow128\rightarrow64$};
\node[note, below=of contextsummary.south] {$h\in\mathbb{R}^{64}$};
\node[note, below=of scorer.south] (scorernote)
  {$3d+64\rightarrow512\rightarrow256\rightarrow1$};
\node[note, below=of softmax.south]        {$u\in\mathbb{R}^K\rightarrow\pi\in\Delta^{K-1}$};

\draw[flow] (candidate)  -- (cardencoder);
\draw[flow] (cardencoder) -- (embedding);
\draw[flow] (pool)       -- (pooler);
\draw[flow] (pooler)     -- (poolsummary);
\draw[flow] (conditions) -- (context);
\draw[flow] (context)    -- (contextsummary);

\draw[flow, rounded corners=2.5pt] (embedding.east)      -- ++(0.55,0) |- (scorer.west);
\draw[flow]                        (poolsummary.east)    -- (scorer.west);
\draw[flow, rounded corners=2.5pt] (contextsummary.east) -- ++(0.55,0) |- (scorer.west);
\fill[black!68] (12.50,-2.40) circle (1.0pt);
\draw[flow] ([yshift=-1.1mm]scorernote.south) -- (softmax.north);
\end{tikzpicture}
\caption{The geometry of DraftFM for one observed pick. Each of the $K$ real
pack cards begins as the same 775-feature representation and becomes one
$d$-dimensional row. The pool is compressed to one $d$-vector, the observed
draft state becomes one 64-vector, and the scorer receives $3d+64$ values for
each candidate. Its $K$ scalar outputs form a probability vector over exactly
the real cards in the pack. Batch and storage padding are omitted because they
do not enter the scorer.}
\label{fig:draftfm-architecture}
\end{figure}

Here $q_j$ is the number of copies of pooled card $j$, $\odot$ is an
elementwise product, and $u_i$ is the candidate's unnormalized score. The same
card encoder and scorer are used for every expansion. The scorer emits one
scalar $u_i$ for each of the $K$ real candidates. Softmax normalization across
those $K$ scores produces a length-$K$ vector of choice probabilities, and the
observed pick identifies the target entry of that vector. Candidates are not
concatenated into a single pack feature vector.

The card encoder applies layer normalization before the first linear layer and
after the final GELU activation. The pool store reserves 46 distinct-card slots.
That figure is an architectural capacity, not an observed count: the largest
pool anywhere in the collection holds 44 distinct cards, and a row that needed
more than the reserved 46 would abort the build rather than be silently
truncated. Each card embedding is augmented by a learned count embedding, with counts
capped at eight. Four learned queries summarize this unordered collection
through four-head cross-attention. This construction does not assign meaning
to the order in which the pool happens to be stored and is closely related to
attention-based models for set-valued inputs \citep{lee2019}. A learned null token
represents the empty P1P1 pool.

The context pathway embeds the drafter's observed game win-rate bucket, number
of games bucket, and event format. 17Lands buckets both covariates for privacy
and does not publicly document their precise computation window, so we use them
as given, as a per-event skill signal rather than a calibrated skill estimate. It also receives seven functions of pack
number, pick number, and current pool size, together with four public shape
descriptors. These are normalized card-list size and indicators for 13-, 14-,
or 15-pick packs. A small multilayer perceptron reduces these inputs to 64 dimensions. The
candidate scorer concatenates the candidate embedding, pool summary, their
elementwise interaction, and this context. A $512$--$256$ multilayer perceptron
then emits one score. A softmax over the valid cards in the pack turns those
scores into choice probabilities. Candidates do not attend directly to one
another in the width-comparison models. Pack composition determines which candidate
scores enter the softmax.

The implementation also permits an attention tower over the complete card list
of a set. A development comparison found that the simpler version without that
tower performed better, so the width comparison keeps that tower off. Every
candidate is still compared with the current pack through the final softmax.
The three widths contain 979,823, 1,637,999, and 3,740,783 fitted parameters,
respectively. None contains a card or set embedding. Their only set-relative
information comes from observable card features, the cards accumulated in the
pool, the cards currently offered, and the four shape descriptors above.

We compared card widths $d\in\{128,256,512\}$ under the fixed recipe of
Table~\ref{tab:fitting-recipe}, evaluated zero-shot on the three development
expansions. Differences were small at the aggregate: the three-set mean top-1
spread was 0.21 percentage points, with set-level variation and no width best on
every set. We selected the middle width, $d=256$, which achieved the best
three-set mean, was at least as accurate as the smaller model on every set
within sampling uncertainty, and uses 44\% of the largest model's parameters and
43\% of its training time. These three expansions are development sets;
Section~\ref{sec:development-design}.
Per-set results for every width and measure appear in
Appendix~\ref{app:additional-results}.

The model used for the prospective forecast is a separate run at that width,
refit from fresh weights on all 32 sets under the same recipe. It carries the
same 1,637,999 fitted parameters as the $d=256$ development model and is
identified by the checkpoint digest \texttt{9442f1de}\ldots
Section~\ref{sec:final-refit-protocol} gives its protocol and provenance.

Training minimizes cross-entropy between the softmax distribution and the
observed 17Lands pick. All picks from a draft remain together. A deterministic
hash of the draft identifier assigns 95\% of drafts from the 29 fitting sets to
training and 5\% to internal validation. This produces 149,483,436 fitting
picks and 7,877,705 reserved validation picks. Internal validation selects the
checkpoint within each run. The internal-validation data exclude all three
development environments. Each check
evaluates at most 200,000 reserved picks distributed approximately evenly
across set-format shards.

\begin{table}[tb]
\centering
\caption{Fixed fitting recipe used for every model width.}
\label{tab:fitting-recipe}
\begin{tabular}{L{0.22\textwidth}L{0.27\textwidth}L{0.41\textwidth}}
\toprule
Component & Setting & Role \\
\midrule
Optimizer & AdamW with $\beta_1=0.9$ and $\beta_2=0.98$ & Adaptive stochastic
optimization with decoupled weight shrinkage \\
\addlinespace
Learning rate & Peak $10^{-3}$ with 2,000-step linear warmup, then cosine decay
to a floor of 1\% of peak &
Stable initial updates followed by progressively smaller steps \\
\addlinespace
Batch construction & 8,192 picks with shard probability proportional to
$n_s^{0.5}$ & Balance large-set coverage against domination by the largest
files \\
\addlinespace
Regularization & Weight decay 0.01, dropout 0.10 in the candidate scorer only,
and label smoothing 0.05 &
Discourage brittle weights and overconfident probabilities \\
\addlinespace
Gradient control & Global gradient norm capped at 1.0 & Prevent an isolated
large update from destabilizing fitting \\
\addlinespace
Stopping rule & At most four corpus-equivalent passes, validation every 2,000
steps, and patience of three checks & Retain the strongest unseen-draft checkpoint
instead of the final update \\
\addlinespace
Random seed & 17 & Make initialization and sampling reproducible \\
\bottomrule
\end{tabular}
\end{table}

These are conventional, fixed optimization controls. The peak learning rate
and weight decay equal the defaults of PyTorch's
\texttt{torch.optim.AdamW}.\footnote{The PyTorch AdamW documentation gives
default values of $10^{-3}$ and 0.01, respectively. See
\url{https://docs.pytorch.org/docs/stable/generated/torch.optim.AdamW.html}.}
The $\beta_2=0.98$ moment coefficient and dropout rate 0.10 follow standard
attention-model practice \citep{vaswani2017}. Weight decay gently
shrinks fitted weights. Dropout randomly suppresses 10\% of the activations in
the candidate scorer's first hidden layer during training and is applied nowhere
else in the network. Label smoothing trains against a mixture of 95\% of the
observed-pick target and 5\% of a uniform distribution over the valid pack.
That last choice is especially natural for behavioral data, where the recorded
selection is an observation rather than a uniquely correct answer.

Every constant in Table~\ref{tab:fitting-recipe} is held fixed across the
width comparison. The sole search dimension is the learned card width
$d\in\{128,256,512\}$. The purpose is to compare representational capacity
under one ordinary, stable recipe, not to optimize a separate training recipe
around each development set. No substantive conclusion depends on treating
these routine constants as uniquely optimal.

Skill conditions the scorer rather than the card representation. All available
players can contribute examples during fitting, while inference can hold the
skill variables at a prespecified deployment value. A P1P1 rating is obtained
by using the empty-pool state and scoring all cards under that same fixed
condition. During an actual draft, the identical model instead scores the cards
in the current pack against the cards already selected.
Section~\ref{sec:experimental-design} will specify how the model width and
deployment condition are chosen.

\section{Development Evaluation}
\label{sec:development-evaluation}

\subsection{Design}
\label{sec:experimental-design}

\subsubsection{Whole-Expansion Development Comparison}
\label{sec:development-design}

The comparison in this paper follows one protocol, fixed in writing before any
comparison number was produced.\footnote{The protocol is versioned as
\texttt{eval-protocol-rebuild-v1}, published as an annotated tag at
\url{https://github.com/brianward92/mtga}, and the evaluation run records that
identifier alongside its results.} Changing a definition in it after
seeing results would invalidate the comparison. An earlier protocol, in which
BRO, TMT, and SOS were development sets and MSH was a frozen single-use test
set, describes a different experiment. That document and its code are preserved
unchanged as a historical artifact and are not used here.

BRO, FDN, and MSH are whole-set development environments. They contributed
neither observed picks nor feature-vocabulary entries to any of the three
candidate models: the card-feature manifest and all three width variants were
fit on the remaining 29 sets. They are inspected, compared across widths, and
used to inform the architecture choice. No set here is a frozen test set, and a
set code is a filter argument and nothing more.

The consequence is stated once, here, and plainly. BRO, FDN, and MSH are
development sets: they selected the model width. Every number this paper reports
on them is development evidence, and it reads optimistic relative to a
never-inspected holdout. The paper has no never-inspected holdout. Later
sections point back to this paragraph instead of restating it.

The evaluation population is every scored pick in the three sets. There is no
sampling, and no skill filter is applied to the population itself.
Table~\ref{tab:dev-population} gives the counts. MSH has no Traditional Draft
shard, so its per-set numbers are Premier Draft only; per-format rows are
reported beside the pooled per-set rows so that this asymmetry is visible rather
than buried in an average. Rows are read from the same memmapped store the
training loop reads. Because these sets were never trained on, the train and
validation split recorded inside a shard carries no meaning here and is not
applied, so all rows are evaluated. Population identity across models is exact:
all three widths score the same rows of the same shards in the same order, which
makes every cross-width comparison paired.

\begin{table}[tb]
\centering
\caption{Evaluation population for the whole-expansion development comparison.
Every scored pick in each set is evaluated, with no sampling.}
\label{tab:dev-population}
\setlength{\tabcolsep}{8pt}
\begin{tabular}{lL{0.24\textwidth}rr}
\toprule
Set & Formats & Picks & Drafts \\
\midrule
BRO & Premier, Traditional & 4,616,619 & 104,676 \\
FDN & Premier, Traditional & 6,346,832 & 152,968 \\
MSH & Premier only & 1,607,786 & 38,580 \\
\midrule
Total & & 12,571,237 & 296,224 \\
\bottomrule
\end{tabular}
\end{table}

Scoring uses the training and serving forward path unchanged. The model runs in
evaluation mode with dropout disabled and gradients switched off. Each pick is
scored from its own observed context, meaning the pool the drafter actually held
and the pack they actually saw; nothing is rolled forward from the model's own
earlier predictions. Conditioning in this comparison is \emph{human mode}: every
pick is scored with the drafter's own observed win-rate and games-played bucket,
exactly as training-time validation does. Deployment-mode conditioning, in which
the skill variables are held at a prespecified value, is a serving question and
belongs to the forecast protocol of Section~\ref{sec:final-refit-protocol}
rather than to this comparison.

Before any pick is scored, the rail verifies and aborts on mismatch that the
on-disk feature manifest's content hash equals the expected value, that every
run record and every checkpoint carries that same manifest hash, and that each
checkpoint file's SHA-256 equals the digest recorded when it was written. A
feature width supplied by a shard that does not match the width a checkpoint
expects is refused rather than silently truncated.

Execution is two-phase, which separates inference from reporting. Phase A runs
each checkpoint once over each set and format and writes one prediction file per
combination, holding one row per scored pick with its rank, probabilities, log
loss, and the drafter's skill fields. Phase B derives every table, curve, and
figure in this paper from those cached files and never loads a network.
Re-deriving a reported number therefore requires no accelerator, and every
number in Section~\ref{sec:development-results} and
Appendix~\ref{app:additional-results} traces to a cached file whose digest is
recorded alongside the run.

Uncertainty is computed at the level of the draft, not the pick, because picks
within one draft are not independent. The rebuild store carries the shard data
but not the original draft identifier, so a draft key is derived from the shard
itself: a draft's picks are contiguous and strictly increasing in pack and pick
number, so a new draft is declared when that key fails to increase or when the
stored split hash of the draft identifier changes. Neither rule can split one
true draft in two, and two adjacent drafts merge only if the pick key increases
across their boundary and they also collide in a 1000-way hash, which is
expected to happen fewer than once per shard against 13,000 to 133,000 drafts.
This is a documented approximation, and it affects only the width of a
confidence interval, never a point estimate.

\subsubsection{Measures and Width Decision}
\label{sec:evaluation-measures}

Let $n$ be the number of real candidates in the pack and let the target rank be
one plus the number of candidates the model scores strictly above the card the
human took. Four measures are reported.

\emph{Top-1 agreement} is the primary behavioral measure. It is the fraction of
picks whose target rank is one, that is, the fraction of picks on which the
model's most probable card is the card the human selected. Using the rank rather
than the argmax means an exact score tie with the human's card counts as
agreement; the two definitions can differ only on exact floating-point ties, and
the rail measures and reports that disagreement count as a diagnostic.
\emph{Top-3 agreement} is the fraction of picks whose target rank is at most
three. \emph{Mean log loss} is the mean of $-\log p$ over picks, in nats, where
$p$ is the probability the softmax over the pack's real candidates assigns to
the card the human took. \emph{Top-label expected calibration error} asks
whether the model's stated confidence in its own recommendation is warranted:
confidence is the probability of the model's argmax candidate, accuracy is
whether that argmax matched the human pick, and the two are compared within 15
equal-mass bins, each holding the same number of picks rather than spanning the
same probability width, and averaged with weights proportional to bin size.

A \emph{forced pick} is a pack with exactly one real candidate.\footnote{A
candidate is a distinct card \emph{name}: duplicate copies of the same card
collapse into one alternative, since choosing either copy is the same choice. A
pack can therefore be forced before its physical cards run out. In FDN, whose
packs are 14 picks long, 1,312 of the 452,845 pick-13 rows, 0.29\%, offer two
physical cards with a single name between them and so count as forced.} It is scored
trivially, because the rank is always one and the probability is always one, so
forced picks inflate top-1 and deflate both log loss and calibration error.
Forced picks are the final pick of each pack almost without exception. We
therefore report every metric twice, once over all picks and once over
non-forced picks only, and treat neither as the sole headline.

The natural reference point for agreement is the \emph{$1/n$ baseline}, the
agreement a uniform-random picker would achieve on the same picks. It is
computed per draft position as the mean of $1/n$ over the picks in that
position, which is a mean of reciprocals rather than the reciprocal of a mean
pack size. Because a pack loses one card per pick, this baseline rises steeply
over a pack and reaches one at the forced final pick, so raw agreement and pack
size move together. Where that mechanical component must be removed we report
chance-normalized agreement, $(\mathrm{acc}-b)/(1-b)$ with $b$ the position's
baseline, excluding forced picks where $b=1$ leaves the quantity undefined.

Each measure is also reported on an \emph{expert} slice as well as on all
drafters. The expert slice is high win rate and high experience together, a
game win-rate bucket of at least 0.55 and a games-played bucket of at least 100,
which is the same definition the earlier protocol used.

All intervals reported in this paper are 95\% percentile cluster bootstraps with
the draft as the resampling unit and $B=1000$ resamples at a fixed seed. Drafts
are drawn with replacement, $n$ from $n$, the statistic is recomputed on the
resampled collection of whole drafts, and the 2.5th and 97.5th percentiles form
the interval. All measures share identical resamples, so their intervals are
mutually consistent, and chance-normalized statistics are recomputed inside each
resample rather than transformed from a raw interval. Calibration error is not a
mean over picks, so it uses a separate kernel in which the bin edges are held
fixed at their full-sample equal-mass values instead of being recomputed in each
resample. That fixed-edge choice is the single approximation in the interval
machinery. The calibration point estimate is unaffected, and the run's self-test
measures the resulting interval discrepancy against the literal estimator on a
subsample and fails if it exceeds $2\times10^{-3}$.

The draft is the resampling unit because picks are not independent. The picks of
one draft share a drafter, an evolving pool, and the packs passed around a pod,
so resampling whole drafts captures arbitrary dependence within a draft by
construction. Independence is then assumed across drafts, not across picks. Two
residual layers are not captured. The same anonymous player contributes multiple
drafts, and player identity is absent from the public data, so player-level
clustering is not possible. Tracked drafters also occasionally share a pod, so a
small fraction of nominally distinct drafts saw the same physical packs. Both
would, if anything, widen the intervals rather than narrow them. At the level of
a single pack-and-pick cell the question does not arise, because each draft
contributes at most one pick there and the cluster bootstrap coincides with the
simple binomial interval.

Results are displayed by set before any average. The unweighted three-set mean
is printed after the per-set rows as a summary and is not a winner rule: a width
that leads on the mean while losing on an individual set must have that reversal
stated explicitly, and equal predictive performance at materially lower size and
runtime is a legitimate reason to prefer a smaller width. The width decision is
made by the author after reviewing predictive performance, parameter count, and
training time together, and the decision rule was fixed before the numbers were
seen.

\subsubsection{Final Refit and Forecast Protocol}
\label{sec:final-refit-protocol}

The model that produces the prospective forecast is not the development
checkpoint under a new name. After the width decision was made, a fresh feature
manifest was built over all 32 sets and a new model was fit at $d=256$ from
randomly initialized weights, with no initialization from any width-comparison
checkpoint. It is a separate run, trained from a clean tree at
repository revision \texttt{7c113e1}. Every constant in
Table~\ref{tab:fitting-recipe} is unchanged, including the random seed 17. The
only differences from a width-comparison run are the corpus, which now includes
BRO, FDN, and MSH, and the feature manifest, which is rebuilt over that corpus.

The corpus is the complete 60-shard collection of 169,932,378 observed picks
described in Section~\ref{sec:data-sources}. The same deterministic split by
draft identifier assigns 161,428,453 picks to fitting and 8,503,925 to internal
validation, and internal validation again selects the checkpoint within the run.
MSH contributes its Premier Draft shard only, because no Traditional Draft file
exists for it. Fitting reached its best internal-validation top-1 agreement of
68.34\% at step 34,000 of 78,822 planned steps, and the run took two hours and
42 minutes of wall-clock time. The fitted model has 1,637,999 parameters, and
its selected checkpoint has SHA-256 digest
\texttt{9442f1de}\ldots. Because these development sets are now inside the
fitting corpus, this internal-validation figure is not comparable to the
zero-shot development numbers in Section~\ref{sec:development-results}, and no
zero-shot claim is made for this model on any set.

The rebuilt manifest has content hash \texttt{793b9db7}\ldots\ and populates its
two data-dependent vocabularies from the 32 fitting sets. The creature-subtype
vocabulary reaches its reserved capacity, filling all 128 slots. The keyword
vocabulary fills 145 of its 166 reserved slots. HOB contributes nothing to
either vocabulary, and this is a verified property rather than an intention: a
control manifest rebuilt with the HOB card records made available to the
vocabulary selector produces a byte-identical content hash, which is only
possible if no HOB card introduced a vocabulary entry. All 321 HOB expansion
records nevertheless encode successfully to the standard 775 finite features, so
no HOB card is missing a representation. HOB introduces exactly one keyword that
the vocabulary does not contain, Mountaincycling, and it is counted in the
unmatched channel described in Section~\ref{sec:card-representation} rather than
receiving a fitted column of its own. HOB therefore contributes public card
descriptions and nothing else. It supplies no picks, no outcomes, no community
ratings, and no fitted vocabulary entries.

Serving artifacts were exported from the selected checkpoint and checked against
the training forward path on real held-out picks. The largest absolute
discrepancy between the training implementation and the exported serving graph
was $3.6\times10^{-6}$, and the two agreed on the recommended card for all 1,536
picks checked. The reported forecast is therefore a property of the fitted
model, not of a serving path that drifted from it.

The forecast condition is a fixed P1P1 query. The format is Premier Draft, the
pool is empty, the position is pack one, pick one, and the pack shape is the
14-pick shape, which is the shape the two most recent sets in our collection use
and the shape HOB uses, since HOB has no bonus sheet. Skill and experience are
held at the threshold pair that defines the expert slice of
Section~\ref{sec:evaluation-measures}, recorded in the seal manifest as win-rate
bucket identifier 28, a game win-rate bucket of at least 0.55, and games-played
bucket identifier 4, a bucket of at least 100 games. The exported serving
artifacts carry a different default, a 0.66 win-rate bucket and the 1{,}000-game
bucket, so the sealed ratings are deliberately not identical to what a deployed
assistant returns under its own defaults. No sensitivity scan over the
conditioning is performed; the condition is chosen once and frozen.

The card universe is every one of the 321 HOB expansion records in the frozen
Scryfall snapshot, which is the complete official card list as that snapshot
mirrors it. Scores are published as a ranking over the 193 unique card names,
identical printings share a score under a documented deduplication rule, and
basic lands are flagged as display-only. This universe is a verified superset of
the 188 non-basic draftable cards that every content-creator source covers, so
no comparison is limited by the model's coverage.

The raw score, rank, and percentile are the statistical forecast. A letter grade
is a presentation layer on top of them, and its rule is frozen before any
comparison statistic is computed. The ladder has 13 levels: every letter from A
to D carries a plus and a minus variant and F is unsigned, giving A+, A, A$-$,
B+, B, B$-$, C+, C, C$-$, D+, D, D$-$, and F. Letters are assigned by fixed
percentile bands of the scored universe, not by fixed score cutoffs, so the
distribution of DraftFM letters is fixed by construction.
Table~\ref{tab:letter-ladder} gives the bands.

\begin{table}[tb]
\centering
\caption{The frozen 13-level letter ladder. Each band is a percentage of the
scored card universe, assigned by rank. The percentages sum to 100.}
\label{tab:letter-ladder}
{\small
\setlength{\tabcolsep}{4.5pt}
\begin{tabular}{lccccccccccccc}
\toprule
Grade & A+ & A & A$-$ & B+ & B & B$-$ & C+ & C & C$-$ & D+ & D & D$-$ & F \\
\midrule
Share (\%) & 2 & 3 & 5 & 8 & 12 & 12 & 13 & 15 & 12 & 8 & 5 & 3 & 2 \\
\bottomrule
\end{tabular}
}
\end{table}

\subsection{Width Comparison Results}
\label{sec:development-results}
\label{sec:whole-expansion-results}

Table~\ref{tab:width-top1} gives the primary result: top-1 agreement on each
development expansion for each of the three card widths, over the full
population of 12,571,237 picks in 296,224 drafts. Per-set rows come first, and
the unweighted three-set mean is printed after them as a summary rather than as
a decision rule. Top-3 agreement, mean log loss, and top-label calibration
error appear in Appendix~\ref{app:additional-results}; the expert-slice and
per-format values accompany the artifact data.

\begin{table}[tb]
\centering
\caption{Top-1 agreement (\%) on the three whole-expansion development sets, all
drafters, formats pooled. The three-set mean is unweighted and is
a summary, not a winner rule. Point estimates are shown; the 95\% percentile
cluster-bootstrap intervals resampling drafts ($B=1000$) have half-widths of at
most $\pm0.09$ percentage points on these values, and complete intervals
accompany the artifact data.}
\label{tab:width-top1}
{\small
\setlength{\tabcolsep}{8pt}
\begin{tabular}{lrrrrr}
\toprule
& & & \multicolumn{3}{c}{Card width $d$} \\
\cmidrule(lr){4-6}
Set & Picks & Drafts & 128 & 256 & 512 \\
\midrule
\multicolumn{6}{l}{\emph{All picks}} \\
BRO & 4,616,619 & 104,676 & 50.63 & 50.83 & 49.89 \\
FDN & 6,346,832 & 152,968 & 60.15 & 60.37 & 60.56 \\
MSH & 1,607,786 & 38,580 & 56.71 & 56.67 & 56.79 \\
\addlinespace
Mean & 12,571,237 & 296,224 & 55.83 & 55.96 & 55.75 \\
\midrule
\multicolumn{6}{l}{\emph{Non-forced picks}} \\
BRO & 4,309,696 & 104,663 & 47.11 & 47.33 & 46.32 \\
FDN & 5,892,768 & 152,941 & 57.08 & 57.32 & 57.52 \\
MSH & 1,492,957 & 38,571 & 53.38 & 53.34 & 53.47 \\
\addlinespace
Mean & 11,695,421 & 296,175 & 52.52 & 52.66 & 52.44 \\
\bottomrule
\end{tabular}
}
\end{table}

The three widths land close together. Across the whole population the largest
per-set difference between any two widths is 0.94 percentage points, on BRO, and
the three-set mean spans only 0.21 points from best to worst. Level differences
between the sets themselves are far larger than differences between widths: BRO
sits near 50\%, MSH near 57\%, and FDN near 60\% at every width. A set's
difficulty, not the model's capacity, dominates this table.

No width wins everywhere, and the reversals are explicit. On BRO the middle
width leads and the largest width is last, 0.94 points behind it with disjoint
intervals. On FDN and MSH the ordering flips and the largest width leads. On FDN
it leads its runner-up, $d=256$, by 0.19 points. On MSH it stands 0.12 points
above $d=256$, but $d=256$ is last there and the runner-up is $d=128$, over
which the lead is 0.08 points with overlapping intervals. The pattern is
identical on non-forced picks
and on the expert slice, so it is not an artifact of the forced-pick rule or of
the population's skill mix. Because the mean-best width is not the set-best
width on two of three sets, the mean is reported as a summary and the reversals
are stated rather than averaged away.

We selected $d=256$. It achieves the best three-set mean on all four
slice-by-policy combinations. Against the smaller width it is at least as
accurate on every set: it leads on BRO and FDN with disjoint intervals, and on
MSH the two are indistinguishable, with intervals that overlap over most of
their length. Against the largest width it gives up 0.19 points on FDN and 0.12
points on MSH and gains 0.94 points on BRO. What it buys is compression.
Table~\ref{tab:width-efficiency} gives the cost side: the selected model holds
1,637,999 parameters, which is 44\% of the largest model's 3,740,783, and it
trained in 3.02 hours against 6.99, or 43\% of the time. It also scores about
2.4 times as many picks per second at evaluation. Paying 2.3 times the
parameters and 2.3 times the training time to move a three-set mean by
$-0.21$ points, while losing nearly a full point on one of the three sets, is
not a trade we take. Equal predictive performance at materially lower size and
runtime was fixed in the protocol as a legitimate reason to prefer a smaller
width, and that is the case here.

\begin{table}[tb]
\centering
\caption{Size and cost of the three widths. Training time is wall clock on the
same device; evaluation throughput is measured over the full 12,571,237-pick
development population. Internal-validation top-1 is measured on reserved drafts
from the 29 fitting sets and is not a development-set result.}
\label{tab:width-efficiency}
{\small
\setlength{\tabcolsep}{7pt}
\begin{tabular}{lrrrrr}
\toprule
Width $d$ & Parameters & Share of $d{=}512$ & Train (h) & Eval (picks/s) &
Internal val.\ top-1 \\
\midrule
128 & 979,823 & 26\% & 2.35 & 230,496 & 68.59\% \\
256 & 1,637,999 & 44\% & 3.02 & 129,494 & 68.56\% \\
512 & 3,740,783 & 100\% & 6.99 & 53,458 & 68.62\% \\
\bottomrule
\end{tabular}
}
\end{table}

Two limits on this table should be carried forward. First, these are development
sets; Section~\ref{sec:development-design}. Second, the internal-validation column of
Table~\ref{tab:width-efficiency} separates the three widths by 0.06 percentage
points and orders them $512>128>256$, which is not the order any width takes on
BRO or FDN. Held-out drafts from sets the model has already seen do not
substitute for a whole unseen expansion, which is the reason the whole-set
comparison exists.

\subsection{Predictive Behavior Across the Draft}
\label{sec:by-position-results}

\begin{figure}[tb]
\centering
\includegraphics[width=\textwidth]{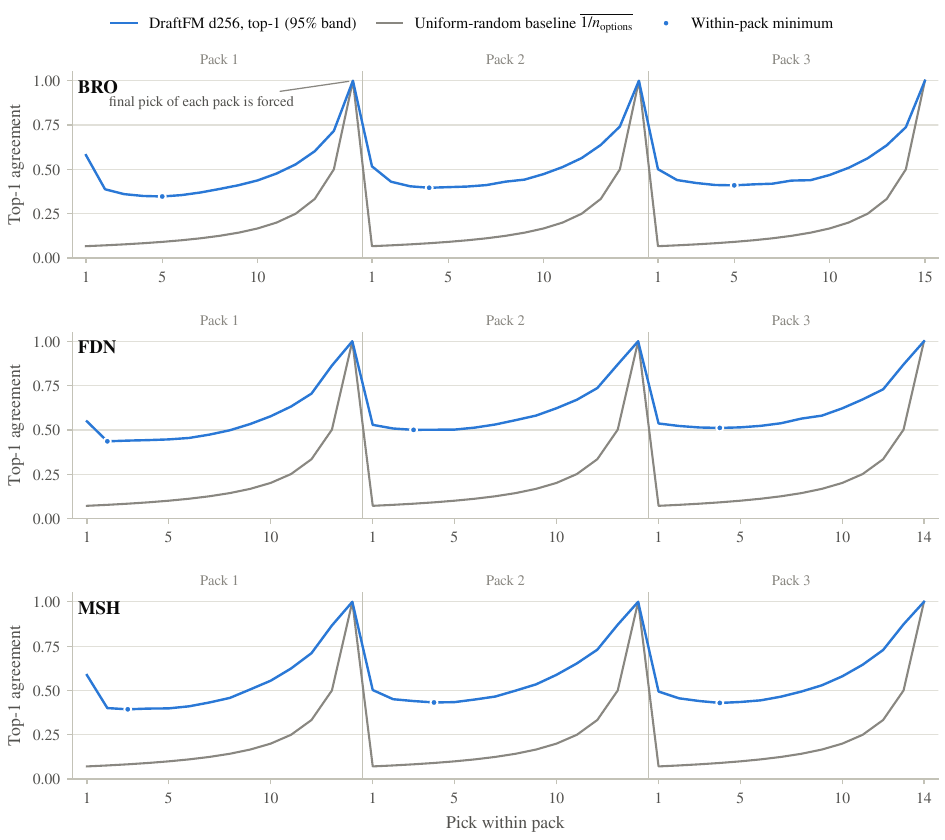}
\caption{Top-1 agreement against draft position on the three development sets,
with the uniform-random baseline, the mean of $1/n$, at each position. DraftFM
$d=256$, all drafters, all picks, formats pooled. Agreement is highest at the
first pick of each pack, falls to a minimum, marked, within the first five picks,
and rises to the end of the pack, where the final pick is forced and agreement is
one by construction. The baseline rises with the same geometry, so pack size and
agreement move together. Shaded bands are 95\% percentile cluster-bootstrap
intervals resampling drafts ($B=1000$); they are narrower than the plotted line
at most positions, with a median width of 0.006.}
\label{fig:top1-by-position}
\end{figure}

\begin{figure}[tb]
\centering
\includegraphics[width=\textwidth]{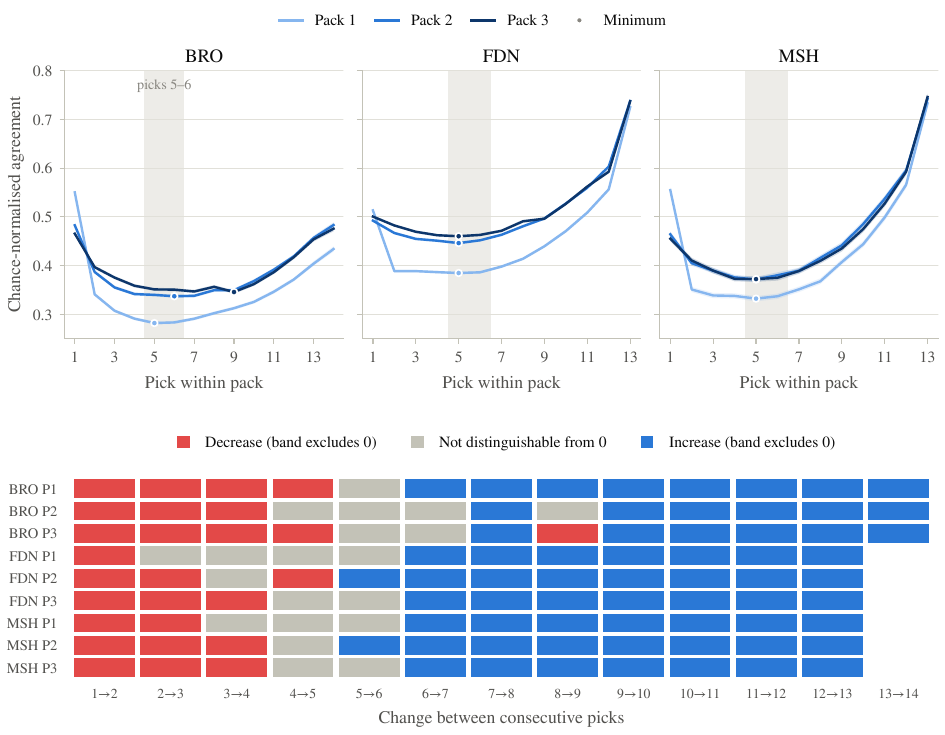}
\caption{Chance-normalized agreement, $(\mathrm{acc}-b)/(1-b)$ with $b$ the mean
of $1/n$ at that position, against pick within pack, by pack. Forced final picks
are excluded because $b=1$ makes the statistic undefined. Top: the curve declines
over the opening picks, turns at pick 5 or 6, shaded, and rises to the end of the
pack. Bottom: the sign of every pick-to-pick change, classified by whether its
95\% interval excludes zero. Bands and step classifications are 95\% percentile
cluster bootstraps resampling drafts ($B=1000$), with the normalized statistic
recomputed inside each resample and $b$ re-derived there. DraftFM $d=256$, all
drafters, formats pooled.}
\label{fig:norm-by-pick}
\end{figure}

Aggregate top-1 agreement conceals a strong positional structure. Following
\citet{ward2021}, we report agreement in each pack-and-pick cell beside the agreement
a uniform-random picker would achieve there, the mean of $1/n$ over the picks in
that cell (Figure~\ref{fig:top1-by-position}). Their pattern reproduces on all
three development sets. Agreement is highest at the first pick of a pack, between
0.49 and 0.59, falls to a minimum within the first five picks, between 0.35 and
0.51, and then climbs to between 0.72 and 0.87 at the last non-forced pick before
the forced final pick trivially reaches one. Because a pack loses one card per
pick, the random baseline climbs over the same span from 0.067 to 0.50. Pack size
and agreement move together, so the aggregate number cannot by itself be read as
evidence about how the model uses context. We therefore also report
chance-normalized agreement, $(\mathrm{acc}-b)/(1-b)$ with $b$ the cell's
baseline, excluding the forced final pick where $b=1$ leaves the quantity
undefined (Figure~\ref{fig:norm-by-pick}).

Normalizing removes the mechanical component but not the positional structure.
The normalized curve declines over the opening picks of every pack, turns at
pick 5 or 6, and rises thereafter, gaining between $+0.125$ and $+0.401$ from
that turn to the last non-forced pick. On eight of the nine pack-by-set curves
the turn is also the curve's lowest point. The exception is BRO pack 3, whose
minimum falls at pick 9, 0.345 against 0.351 at pick 5 and 0.350 at pick 6, and
that is where Figure~\ref{fig:norm-by-pick} places that curve's minimum marker.
Measured from the true minima rather than from the turn, the gains run from
$+0.130$ to $+0.401$. We tested whether the rise is monotone from an early pick,
as an earlier reading of these curves had suggested, and it is not. Taking pick
3 as the start, none of the nine curves is monotone and ten pick-to-pick
decreases have 95\% intervals excluding zero; from pick 4, four remain, three of
them at the step into pick 5. The turn itself is clean: at picks 5 and 6 no
curve shows a band-supported decrease, and from pick 6 onward the increase is
band-supported at essentially every step. The sole exception after pick 5 is
that same BRO pack 3 step from pick 8 to 9, a decrease of 0.011 with a 95\%
interval of $[-0.016,-0.006]$. The rise also does not generally restore the
pack's opening level. It does not in BRO packs 1 and 2: pack 1 never returns to
its pick-1 value, and pack 2's final pick clears that value by 0.0005, far
inside the intervals. It does in BRO pack 3, where pick 14 reaches 0.476
$[0.470,0.481]$ against 0.465 $[0.462,0.469]$ at pick 1, and in FDN and MSH,
where the pick-1 value is exceeded from pick 9 to 12 onward. The description we
carry forward is therefore that normalized agreement declines over the first
four or five picks, turns at pick 5 or 6, and rises at essentially every later
step. These are
descriptive statements about the fitted policy's predictions. Nothing here
attributes the pattern to the pool encoder or to any other component, which would
require an identifying comparison this design does not supply.

Two further positional results are reported in
Appendix~\ref{app:additional-results} rather than here. Picks by experienced,
high-win-rate drafters are harder to predict than the population, and that gap is
concentrated at the very beginning of the draft, reaching $-2.8$, $-3.6$, and
$-5.3$ normalized percentage points at the first pick on BRO, FDN, and MSH before
becoming indistinguishable from zero across most of pack 3
(Figure~\ref{fig:expert-gap}). Over the same span the model is overconfident
everywhere and progressively less so, with mean top-label confidence exceeding
top-1 agreement by 0.068 to 0.078 in pack 1 and by 0.032 to 0.057 in pack 3
(Figure~\ref{fig:calibration-by-position}).

The by-position shape is a property of the model family rather than of the
selected width. Paired differences between widths on the normalized curve are
small against a curve that spans 0.28 to 0.75, and the between-width Pearson
correlation of the curves is at least 0.996, so the shape is width-invariant even
where a level difference is resolvable. Those paired differences are bootstrapped
directly, since all three widths scored identical pick populations.

\section{Prospective HOB Forecast}
\label{sec:hob-forecast}

\subsection{Forecast and Evaluation Timeline}
\label{sec:hob-timeline}

The release of a \emph{Magic: The Gathering} set occurs over several stages rather than on a single date. Wizards of the Coast begins by previewing individual cards through official articles and participating content creators. It then publishes a complete card image gallery before the set becomes widely available for play. This creates a short interval during which the complete public card list is available but no large-scale digital draft data yet exist.

For \emph{Magic: The Gathering | The Hobbit} (HOB), the preview season began on July 18, 2026, and the complete card image gallery became available on July 31, 2026.\footnote{\url{https://magic.wizards.com/en/news/feature/where-to-find-the-hobbit-previews}} The set was first playable at pre-Prerelease events at Gen Con from July 30 through August 2. Local game store (LGS) Prerelease events began on August 7, HOB was scheduled to become available on MTG Arena on August 11, and its global tabletop release was scheduled for August 14.\footnote{\url{https://magic.wizards.com/en/news/feature/where-to-play-the-hobbit}} Table~\ref{tab:hob-release-timeline} summarizes this chronology. Date ranges are ordered first by their starting date and then by their ending date.

\begin{table}[tb]
\centering
\caption{Release chronology for \emph{Magic: The Gathering | The Hobbit}.}
\label{tab:hob-release-timeline}
\setlength{\tabcolsep}{9pt}
\begin{tabular}{lll}
\toprule
Event & Start & End \\
\midrule
Official preview season & 2026-07-18 & 2026-07-31 \\
Gen Con pre-Prerelease events & 2026-07-30 & 2026-08-02 \\
Complete card image gallery & 2026-07-31 & 2026-07-31 \\
Local game store Prerelease events & 2026-08-07 & 2026-08-13 \\
MTG Arena release & 2026-08-11 & 2026-08-11 \\
Global tabletop release & 2026-08-14 & 2026-08-14 \\
\bottomrule
\end{tabular}
\end{table}

During this interval, Limited-focused content creators publish set reviews in which they evaluate the cards before acquiring substantial experience with the completed draft format. These reviews commonly assign ordinal letter grades, sometimes including plus and minus modifiers, grade ranges, or a separate ``build-around'' designation.\footnote{A build-around card may be weak in a typical deck but become valuable when the drafter obtains the appropriate supporting cards.} Because a review may be distributed across several videos or podcast episodes, we record both the first and last publication dates represented by each source. Table~\ref{tab:hob-expert-sources} records the HOB material available as of August 9, 2026. Sources are ordered by the left endpoint of their publication interval and then by the right endpoint.

\begin{table}[tb]
\centering
\caption{Pre-Arena expert HOB set-review sources available as of August 9,
2026.\,*Retrieved date; the source carries no publication date.}
\label{tab:hob-expert-sources}
{\small
\setlength{\tabcolsep}{3pt}
\begin{tabular}{L{0.17\textwidth}L{0.215\textwidth}L{0.15\textwidth}
                L{0.225\textwidth}L{0.13\textwidth}}
\toprule
Source & Rater or raters & Medium & Publication interval & Coverage \\
\midrule
Nizzahon Magic & Nizzahon & YouTube & 2026-07-31--2026-08-03 & Full set in six videos \\
\addlinespace
Card Game Base & An\v{z}e Mlakar & Website & 2026-08-01 & Full draftable set (188 cards) \\
\addlinespace
Draftsim & Andrew Quinn & Website & 2026-08-03 & Full draftable set (188 cards) \\
\addlinespace
Limited Resources & Marshall Sutcliffe and Luis Scott-Vargas & Podcast & 2026-08-03 & Commons and uncommons \\
\addlinespace
Limited Level-Ups & Marc Anderson & YouTube and website & 2026-08-05 & Full draftable set (188 cards) \\
\addlinespace
NicolaiBolas & NicolaiBolas & 17lands tier list (via patron link) & 2026-08-09* & Full draftable set (188 cards) \\
\bottomrule
\end{tabular}
}
\end{table}

Nizzahon's six-part review covered multicolored and colorless cards followed by the five colors in WUBRG order. The series began on July 31 and concluded on August 3.\footnote{The interval is anchored by the original first and final videos. See \url{https://www.youtube.com/watch?v=s_B_p3sOg8c} and \url{https://www.youtube.com/watch?v=rRGwwbK3Tlo}.} Card Game Base published the earliest full-coverage written tier list, a single-author A-to-F grading of all 188 draftable non-basic cards by An\v{z}e Mlakar on August 1.\footnote{\url{https://cardgamebase.com/the-hobbit-draft-tier-list/}, \texttt{datePublished} 2026-08-01.} Draftsim's set review by Andrew Quinn, published August 3, graded the same 188-card universe on a 0--10 numeric scale;\footnote{\url{https://draftsim.com/mtg-hob-limited-set-review/}, \texttt{datePublished} 2026-08-03.} Limited Resources episode 865, published August 3, covered every HOB common and uncommon.\footnote{\url{https://lrcast.com/limited-resources-865-the-hobbit-set-review-commons-and-uncommons/}} Limited Level-Ups released a three-part video guide on August 5--6 with a companion tier list, maintained by co-host Marc Anderson, whose pre-Prerelease state (last updated 2026-08-05) grades the full 188-card universe.\footnote{\url{https://limitedlevelups.com/tier-list}; the list is served from a live endpoint and may change after release, so we preserve the snapshot retrieved on 2026-08-09 reflecting its 2026-08-05 state.} As of the August 9 cutoff, the Limited Resources feed contained no corresponding HOB rare-and-mythic episode.\footnote{The original publication feed checked at the cutoff was \url{https://lrcast.com/feed/}.} NicolaiBolas's full per-card tier list was transcribed on August 9 from his 17lands tier-list export and is included in the comparisons below; his file carries a retrieval date rather than a publication date, and republication permission for the complete grade table is still pending, so we report agreement statistics rather than his grades. If a source later provides separate ratings from multiple hosts, those ratings will be preserved separately. If the hosts publish a single consensus grade, the podcast will be treated as one forecast source.

The letter systems these creators use are informal but remarkably consistent conventions rather than ad hoc labels. Introducing the Limited Resources review, Luis Scott-Vargas described the scale as ``A through F with two subgrades,'' where the A's are ``the bombs and game-winners, cards that are good in many situations, especially when behind,'' the B's ``actively pull you towards their colors,'' the C's are playable and interchangeable---``they're the pawns of Limited''---the D's are cards ``you're unhappy to run,'' and the F's are ``basically just straight up unplayable''; the two subgrades mark sideboard cards and build-arounds, ``cards that by themselves don't do anything, but in the right deck, they can be your best card.''\footnote{Limited Resources episode 865, at approximately 00:11:16--00:13:26; quoted from the episode's automated transcript with punctuation regularized.} This shared vocabulary is what makes cross-creator comparison meaningful: every source in Table~\ref{tab:hob-expert-sources} publishes per-card ordinal evaluations on either this letter convention or an explicit numeric scale, and each was released before large-scale play of the completed format was possible.

We use the interval between publication of the complete gallery and the August
11 MTG Arena release to establish a verifiable pre-Arena forecast milestone.

That milestone has been met. The forecast was generated at 2026-08-09T23:33:45Z
and published at \url{https://github.com/brianward92/draftfm} under the annotated
tag \texttt{draftfm-v1.0}, whose server-side tagger date is 2026-08-09T23:56:16Z,
with the accompanying release published one second later. Both timestamps precede
the scheduled August 11 MTG Arena release by roughly 36 hours.

The published artifact contains scores, ranks, percentiles, and the frozen
presentation grades for all 321 HOB expansion records, together with the manifest
that identifies everything used to produce them.
Table~\ref{tab:seal-artifact} gives the SHA-256 digest of each published file.
The forecast was produced by checkpoint \texttt{9442f1de}\ldots\ under feature
manifest \texttt{793b9db7}\ldots, trained at repository revision \texttt{7c113e1}
and scored at revision \texttt{eb27b8c}; the manifest also records the card-list
and printing-list digests and the full Scryfall source chain, and
Appendix~\ref{app:reproducibility} reproduces that chain.

\begin{table}[tb]
\centering
\caption{Files in the published seal and their SHA-256 digests. Each file was
re-downloaded from the published tag and re-hashed; every digest matched the
value recorded at generation time.}
\label{tab:seal-artifact}
{\scriptsize
\setlength{\tabcolsep}{4pt}
\begin{tabular}{L{0.30\textwidth}L{0.31\textwidth}L{0.30\textwidth}}
\toprule
File & SHA-256 & Contents \\
\midrule
\path{hob_p1p1_forecast.csv} &
\texttt{a16c34767aa382e7\allowbreak 703f480207f84804\allowbreak 84e29ad22bbab214\allowbreak 2aa2195e9b9a94a9} &
Rank, name, score, percentile, and grade for all 321 records \\
\addlinespace
\path{hob_p1p1_forecast.parquet} &
\texttt{4c539b9cd78d669c\allowbreak 8b9d9b79745d8009\allowbreak f0fc65dc592cc600\allowbreak 7bbba91767e20e25} &
The same table in columnar form \\
\addlinespace
\path{seal_manifest.json} &
\texttt{f7832e79f918bd5b\allowbreak 0f04daf0c280a264\allowbreak 4941a78f7a74a4a1\allowbreak b7496924b9d7b553} &
Conditioning, universe, ladder definition, and the full provenance chain \\
\addlinespace
\path{README.md} &
\texttt{a6fa0399cb0be38f\allowbreak 0a9ebdd3495a53fc\allowbreak 2692e57f7e552b00\allowbreak bfdc00cb1d54e48c} &
Plain-language description of the artifact and its limits \\
\bottomrule
\end{tabular}
}
\end{table}

Two properties make the seal checkable rather than merely asserted. Scoring is
CPU-only and reads no clock, hostname, or random number generator, and it was
executed four times in total, twice under the committed scoring revision, with
byte-identical CSV and Parquet output on every run. Independently of that, all
four published files were downloaded again from the tag after publication and
re-hashed, and each digest matched.

The milestone is specifically a \emph{pre-Arena} forecast. It is not a
claim that no person had previously played with HOB cards. Limited human play
had already occurred at Gen Con and at local game store Prerelease events. The
timestamps establish the narrower claim that the reported DraftFM ratings
existed before HOB became available for large-scale drafting on MTG Arena.

The prospective artifact does not use HOB draft outcomes from 17Lands and does
not claim that DraftFM is more accurate than any expert reviewer. The
out-of-sample outcome evaluation begins when 17Lands first publishes a
sufficiently complete HOB draft dataset. We will freeze that dataset by
retrieval time and cryptographic hash and publish a follow-on note comparing the
sealed DraftFM predictions head-to-head with the pre-Arena expert forecasts.
Every forecast will be evaluated on the same set of matched cards and against
the same realized HOB outcomes. The follow-on comparison will be reported
regardless of which forecast performs best.

\subsection{Comparison with Content Creators}
\label{sec:creator-comparison}

As the model is intended to be a pick predictor, we obtain an abstract card rating by scoring every card under the fixed P1P1 conditions defined above. The raw numerical score, rank, and percentile constitute the statistical forecast. A score-to-letter mapping, frozen before examining the HOB comparisons, is included only as a presentation layer that makes the model output readable alongside the content creators' grades.

The deployment conditioning that defines those P1P1 conditions, and the
score-to-letter rule that turns a score into a grade, are both stated in
Section~\ref{sec:final-refit-protocol} and were frozen before any comparison
statistic was computed.

Comparing sources requires one common scale. Every source is placed on the same
13-level letter ladder used for DraftFM's own grades, running A+, A, A$-$, B+,
B, B$-$, C+, C, C$-$, D+, D, D$-$, and F. Sources that already publish letters
map directly onto it, and a plus or a minus is simply its own level rather than a
decoration on the letter beside it. A source that publishes a numeric scale is
converted by rank: its cards are ordered by their published numbers and cut at
the same fixed band percentages that produce DraftFM's letters
(Table~\ref{tab:letter-ladder}), so no comparison depends on how one reviewer's
0-to-10 scale happens to line up with another's letters. Where a source publishes
a numeric score, rank statistics use the raw numbers rather than the letters
derived from them, which avoids discarding resolution the source actually
provided.

Two mechanical consequences of that conversion should be read with the
step-based statistics. DraftFM's own letters are apportioned to the frozen
percentages by largest remainder, while a numeric source is cut at the
cumulative band boundaries. The target percentages are identical, but the two
rules can realize a given band with one more or one fewer card. Separately, a
coarse integer scale cannot always realize the bands at all: Draftsim's review
puts 47 of its 188 cards on the single grade 4, a quarter of the set and wider
than any band, which leaves its C+ and D$-$ bands holding no cards. Draftsim's
exact-match and within-one-step rates are therefore coarse projections of a
scale with fewer distinguishable levels than the ladder has rungs. The rank
statistics use the raw numbers and are unaffected by both points.

Three rules govern awkward grades, and all three were fixed before any statistic
was computed. A grade written as a range or a pair contributes its primary grade
only. A separate build-around or sideboard designation is recorded with the card
but is not used in any statistic, because it describes a condition for playing
the card rather than a position in the ordering.\footnote{This rule is why
Limited Level-Ups, whose tier list covers the full 188-card draftable set,
contributes $n=187$ to every pair it appears in. It designates one card, The
Black Arrow, ``SB'' for sideboard-only, with no ordinal grade beside it, so that
card carries no position in its ordering and is dropped. The Black Arrow is an
uncommon, so it also falls inside the 120 commons and uncommons Limited
Resources covers, which is why that particular pair is computed on 119 shared
cards rather than 120.} A card a source did not grade is
missing, not bad: it is dropped from every pair involving that source and is
never imputed, and in particular an ungraded rare is never treated as a low
grade. Sources differ in coverage, most visibly where a review covers commons and
uncommons but not rares and mythics, so every statistic is computed on the cards
the two sources share and the number of shared cards is reported with it.

On that common footing we report five quantities for each pair of sources,
ordered from the most directly interpretable to the least. First, the
\emph{exact match rate}: the share of shared cards on which the two sources
assign the identical grade. Second, the \emph{within-one-step rate}: the share on
which they are identical or one rung apart, so that B+ against B counts as
agreement and B+ against B$-$ does not. Third, the \emph{rank correlation}: how
closely the two orderings track one another overall, computed as the
tie-corrected Spearman coefficient. Fourth, a \emph{rank agreement} coefficient
that counts the pairs of cards the two sources order the same way against the
pairs they order oppositely, with a correction for the many ties letter grades
produce.\footnote{Kendall's $\tau_b$.} Fifth, \emph{top-10 overlap}: the number
of cards appearing in both sources' top ten. Because letters tie heavily at the
top, a source's top ten is taken as every card at or above the grade of its tenth
card, which can hold more than ten cards; the reported figure is the size of the
intersection.

These are measures of agreement among pre-Arena forecasts. None of them is a
measure of predictive accuracy, and none can become one before HOB outcome data
exist.

Table~\ref{tab:creator-agreement} reports these quantities for every pair of
human sources, and then for DraftFM against each of them. The DraftFM rows were
computed only after the forecast was sealed and published, so no comparison
statistic could have influenced the model output.

\begin{table}[tb]
\centering
\caption{Pairwise agreement among the pre-Arena forecasts, human and model, on
the common 13-level ladder. Statistics use only the cards a pair shares, and $n$
is that count. Rows are ordered by exact match rate within each block; the lower
block holds the sealed DraftFM forecast against each human source. Two notes on
construction. The sealed letters were banded over all 193 HOB card names, basic
lands included, and all five basics landed in D, so on the shared 188-card
comparison set DraftFM's D band holds 5 cards, 2.7\% against the 5\% the band
targets; that is an artifact of banding a universe wider than the comparison
set, not a defect in the forecast. And a source's top ten is inclusive of every
card tied with its tenth, so the intersected sets are not all the same size:
DraftFM is the only tie-free source and contributes exactly ten cards, while the
creators' inclusive sets run as large as 27, for Limited Resources. The overlap
column is not a measure of selectivity.}
\label{tab:creator-agreement}
{\small
\setlength{\tabcolsep}{6pt}
\begin{tabular}{L{0.375\textwidth}rrrrrr}
\toprule
Pair & $n$ & Exact & Within one & Spearman & Rank agr. & Top-10 \\
\midrule
Limited Resources vs.\ Nizzahon Magic & 120 & 0.308 & 0.683 & 0.583 & 0.497 & 8 \\
Card Game Base vs.\ NicolaiBolas & 188 & 0.303 & 0.723 & 0.748 & 0.631 & 11 \\
Card Game Base vs.\ Nizzahon Magic & 188 & 0.303 & 0.569 & 0.684 & 0.577 & 12 \\
Card Game Base vs.\ Limited Level-Ups & 187 & 0.294 & 0.668 & 0.785 & 0.667 & 10 \\
Draftsim vs.\ Limited Resources & 120 & 0.283 & 0.617 & 0.610 & 0.501 & 14 \\
Limited Resources vs.\ NicolaiBolas & 120 & 0.275 & 0.675 & 0.570 & 0.472 & 14 \\
Card Game Base vs.\ Draftsim & 188 & 0.271 & 0.575 & 0.675 & 0.565 & 10 \\
Card Game Base vs.\ Limited Resources & 120 & 0.250 & 0.642 & 0.494 & 0.409 & 8 \\
Limited Level-Ups vs.\ Limited Resources & 119 & 0.244 & 0.605 & 0.536 & 0.430 & 8 \\
Draftsim vs.\ Limited Level-Ups & 187 & 0.241 & 0.599 & 0.678 & 0.555 & 5 \\
Limited Level-Ups vs.\ Nizzahon Magic & 187 & 0.235 & 0.578 & 0.664 & 0.554 & 8 \\
Draftsim vs.\ NicolaiBolas & 188 & 0.234 & 0.628 & 0.693 & 0.594 & 7 \\
NicolaiBolas vs.\ Nizzahon Magic & 188 & 0.234 & 0.606 & 0.744 & 0.623 & 9 \\
Limited Level-Ups vs.\ NicolaiBolas & 187 & 0.198 & 0.620 & 0.641 & 0.535 & 8 \\
Draftsim vs.\ Nizzahon Magic & 188 & 0.175 & 0.569 & 0.698 & 0.590 & 7 \\
\midrule
DraftFM vs.\ Nizzahon Magic & 188 & 0.261 & 0.532 & 0.635 & 0.495 & 6 \\
Card Game Base vs.\ DraftFM & 188 & 0.186 & 0.415 & 0.437 & 0.320 & 7 \\
DraftFM vs.\ Draftsim & 188 & 0.181 & 0.484 & 0.581 & 0.439 & 5 \\
DraftFM vs.\ NicolaiBolas & 188 & 0.175 & 0.479 & 0.596 & 0.446 & 3 \\
DraftFM vs.\ Limited Resources & 120 & 0.150 & 0.583 & 0.594 & 0.461 & 6 \\
DraftFM vs.\ Limited Level-Ups & 187 & 0.118 & 0.455 & 0.399 & 0.289 & 3 \\
\bottomrule
\end{tabular}
}
\end{table}

The spread in that table is the context DraftFM will be read against. No two
human reviewers assign the same grade to even a third of the cards they both
covered, and the best-agreeing pair matches exactly on 30.8\% of shared cards.
Allowing one rung of slack raises agreement to between 0.57 and 0.72, and the
ordering statistics sit between 0.49 and 0.79 for rank correlation and between
0.41 and 0.67 for rank agreement. Expert forecasts of a new Limited format
disagree substantially with one another before the format is played, which is the
baseline any single forecast, including ours, should be judged against.

Against that baseline, DraftFM sits inside the human spread on the ordering
measures and outside it on the step measures. Its rank agreement with four of the
six creators falls within the range the creators span with one another, though
toward the lower part of it, and it agrees most with Nizzahon Magic at 0.495. No
DraftFM pair exceeds the creator maximum on any of the five statistics. The
clearest separation is the within-one-step rate, where five of the six DraftFM
pairs fall below the lowest value any creator pair reaches: when DraftFM and a
creator disagree, the disagreement tends to land farther down the ladder than
creators' disagreements land from each other. These remain measures of agreement
among forecasts, and none of them says which forecast is right.

In the follow-on paper, once 17Lands data are available, DraftFM can be evaluated in its natural pick-prediction domain. To compare it with a content creator, we induce a static pick policy from that creator's letter grades. The policy selects a card with the highest grade in the observed pack. If $N$ cards share the highest grade and the human drafter selects one of them, the creator policy receives $1/N$ expected agreement for that pick. Equivalently, the policy chooses uniformly among all tied highest-grade cards.

This rule can be applied to any pick for which the creator graded every candidate. DraftFM, by contrast, is context-dependent and receives features describing the pool drafted so far. We therefore hypothesize that static creator policies will lose agreement relative to DraftFM after P1P1 because they cannot observe color commitment, card synergy, or the existing pool. P1P1 and later-pick results will be reported separately. We treat this comparison as a stylized horse race. The creator grades were published as general card evaluations, not as fully context-dependent draft policies.

\subsection{Five Cards the Model Likes}
\label{sec:five-cards}

\newcommand{\hobcard}[5]{%
  \begin{minipage}[t]{0.30\textwidth}
  \centering
  \includegraphics[width=\linewidth]{figs/cards/#1}\\[3pt]
  {\footnotesize #2.\ #3\\ Score #4 \quad Grade #5\par}
  \end{minipage}}

\begin{figure}[tb]
\centering
\hobcard{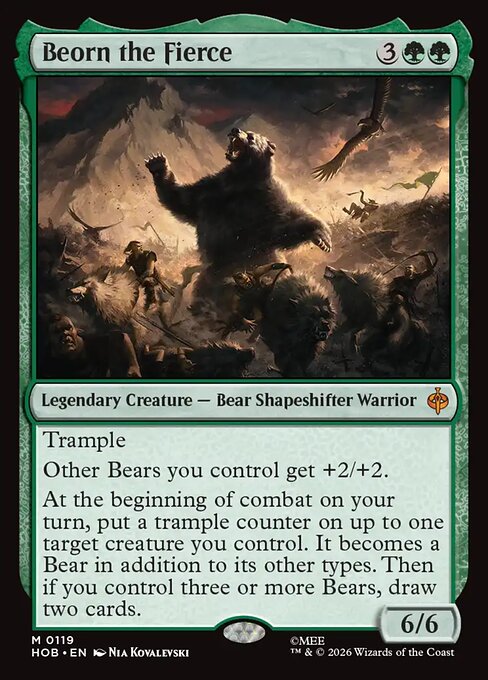}{1}{Beorn the Fierce}{5.011}{A+}\hfill
\hobcard{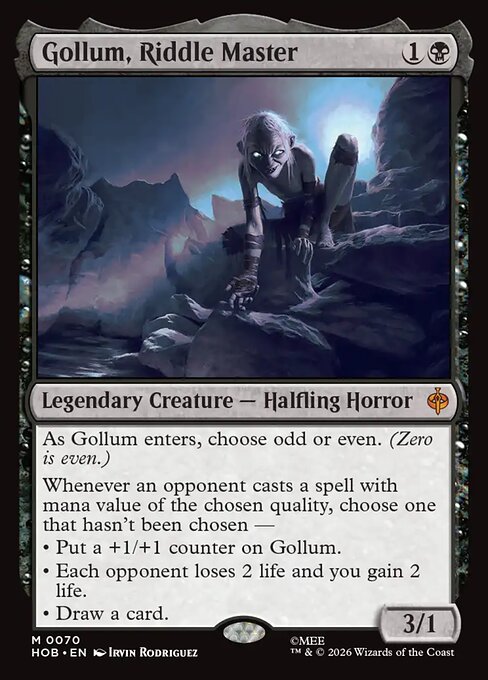}{2}{Gollum, Riddle Master}{4.425}{A+}\hfill
\hobcard{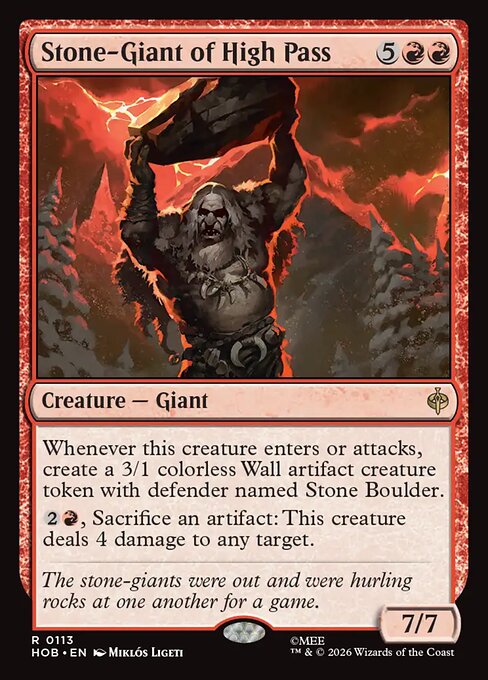}{3}{Stone-Giant of High Pass}{4.393}{A+}

\vspace{12pt}

\hobcard{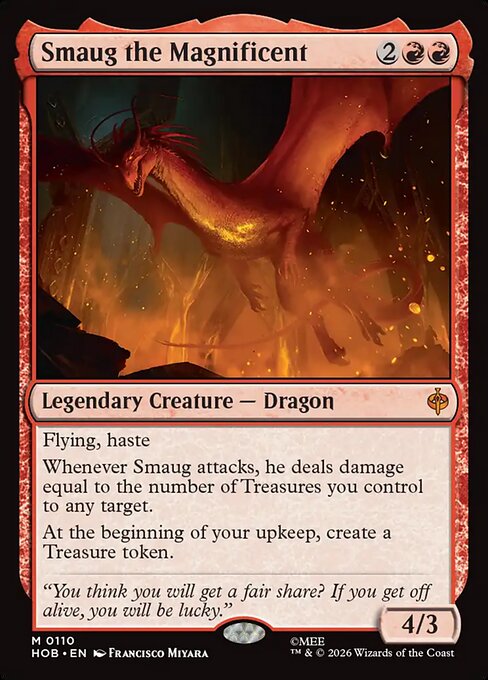}{4}{Smaug the Magnificent}{4.313}{A+}\hspace{0.04\textwidth}
\hobcard{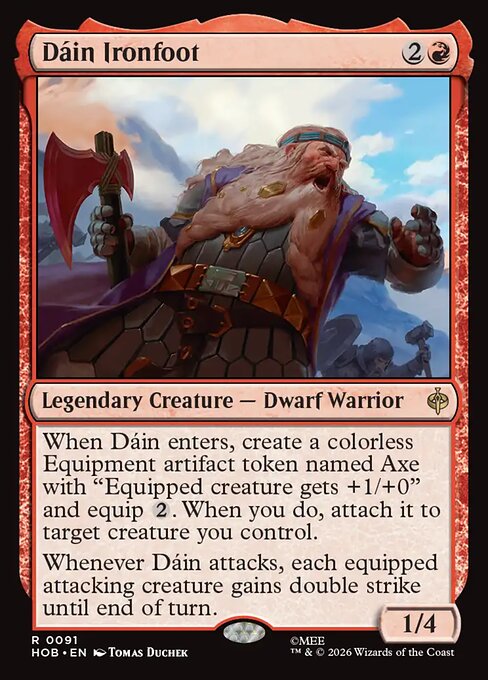}{5}{D\'ain Ironfoot}{4.116}{A}

\caption{The five highest-rated HOB cards in the sealed DraftFM forecast. Scores
are the model's raw P1P1 scores under the frozen conditioning of
Section~\ref{sec:final-refit-protocol}, and grades come from the fixed
percentile bands of Table~\ref{tab:letter-ladder}. Both are taken from the sealed
artifact and were fixed before the set was playable on MTG Arena. Card images are
copyright Wizards of the Coast and are reproduced here at reduced size for
scholarly commentary; images are served by Scryfall.}
\label{fig:hob-top-five}
\end{figure}

In my opinion, these line up with my expectations as great cards. Beorn is a
must-kill if you have other creatures and with one or two other bears in your
deck, the draw-two will occur often enough. Gollum is a great two-drop that
gives great value. And finally ranks 3 through 5 suggest red will be very
strong and this combination of rares in a sealed pool would be excellent.

\section{Conclusion}
\label{sec:conclusion}

DraftFM demonstrates that a single feature-based policy, fitted across many
draft environments, predicts human draft choices in whole expansions that
contributed nothing to its training. On three expansions withheld in their
entirety, one 1.6-million-parameter network reaches 50.8\%, 60.4\%, and
56.7\% top-1 agreement, and the positional structure of that agreement
reproduces and sharpens the pattern first reported for single-set models:
raw agreement is governed by pack geometry, and once chance is removed the
curve declines over the opening picks, turns at pick 5 or 6, and rises at
essentially every later step. The model is uniformly overconfident, less so
as drafts progress, and the picks of expert drafters are hardest to predict
exactly where a set review matters most, at the first pick of the draft.

The second contribution is procedural. The HOB forecast was scored under
frozen conditions, exported byte-identically, hashed, and published with its
complete provenance chain roughly 36 hours before the set became draftable
on MTG Arena. Against six independent pre-release reviewers, the sealed
ranking agrees with each of them roughly as much as reviewers agree with one
another on most measures, sitting toward the lower part of that range and
never above it, and its near-misses land farther down the grade ladder than
theirs do. The paper claims no predictive accuracy for the forecast. Before
outcome data exist, agreement among forecasts is all that can be measured,
and it is reported as exactly that.

The commitment stands regardless of outcome. When 17Lands publishes a
sufficiently complete HOB draft dataset, we will freeze it by retrieval time
and cryptographic hash and report the evaluation of every pre-Arena
forecast, ours and the reviewers', on the same matched cards and the same
realized outcomes. Day zero is not a corner case; it is the permanent
condition of any model deployed into an environment that keeps inventing
new pieces. Drafting merely makes it measurable.

\section*{Acknowledgments}

Data from 17Lands.com (CC BY 4.0) and card data from Scryfall. Many thanks to
the content-creators whose public pre-release evaluations are analyzed in
Section~\ref{sec:hob-forecast}: Nizzahon, Andrew Quinn, An\v{z}e Mlakar,
Marc Anderson, Marshall Sutcliffe, Luis Scott-Vargas, and NicolaiBolas. The
software implementation, statistical analyses, and manuscript preparation
were carried out with assistance from AI tools, OpenAI's Codex and
Anthropic's Claude Code. All scientific decisions and the final prose are
the author's, and responsibility for the results rests with the author
alone. \emph{Magic: The Gathering} and its card images are property of
Wizards of the Coast LLC; this independent research is not affiliated with
or endorsed by Wizards of the Coast.

\medskip
\noindent I would also like to thank my wife, Sara, who has supported me on every
journey in life, no matter how big or small.

\appendix

\section{Public Artifact and Reproducibility Record}
\label{app:reproducibility}

The sealed forecast, its manifest, and a plain-language README are published at
\url{https://github.com/brianward92/draftfm} under the annotated tag
\texttt{draftfm-v1.0}, whose server-side tagger date is 2026-08-09T23:56:16Z,
with the accompanying release published one second later. The manifest, SHA-256
\texttt{f7832e79}\ldots, is the authoritative reproducibility record: it carries
the complete hash chain from the raw Scryfall bulk snapshot through the processed
card tables, the feature manifest, the model checkpoint, the training and scoring
code revisions, and the digests of both output files, together with the
conditioning, the letter-ladder definition, and the determinism records. Readers
who want to verify any link in that chain should read the published manifest
rather than this appendix, which prints only the digests the paper itself cites:
raw Scryfall snapshot \texttt{4a60c20e}\ldots, feature manifest
\texttt{793b9db7}\ldots, model checkpoint \texttt{9442f1de}\ldots, and the
manifest above. Table~\ref{tab:seal-artifact} gives the published files and their
full digests.

The pipeline runs in seven stages: acquire the 17Lands draft files and a dated
Scryfall snapshot and freeze their hashes; curate the wide source files into one
canonical columnar schema; build the frozen card feature table and its manifest,
holding out from vocabulary fitting any set the experiment must not see; shard
the picks into memory-mapped training stores; fit; evaluate under the versioned
protocol; and export and seal. Software versions were Python 3.12.13, PyTorch
2.12.1, and ONNX Runtime 1.28.0, with the frozen sentence encoder run once in a
separate environment on PyTorch 2.13.0 and sentence-transformers 5.7.0. Training
and development evaluation used an Apple Silicon GPU; export and forecast scoring
are CPU-only.

Two checks support the seal. Serving artifacts exported from the selected
checkpoint were compared against the training forward path on real held-out
picks: the largest absolute discrepancy was $3.6\times10^{-6}$, and the two
agreed on the recommended card for all 1,536 picks checked, so the reported
forecast is a property of the fitted model and not of a serving path that
drifted from it. Separately, forecast scoring reads no clock, hostname, or random
number generator; it was executed four times in total, twice under the committed
scoring revision, and produced byte-identical CSV and Parquet output on every
run. After publication all four files were downloaded again from the tag and
re-hashed, and every digest matched.

\section{Additional Results}
\label{app:additional-results}

\subsection{Complete Development Results}
\label{app:full-dev-tables}

Table~\ref{app:res-all} gives every measure the rebuild protocol defines, for
every width, on all three development sets, under both forced-pick policies, with
formats pooled. Section~\ref{sec:whole-expansion-results} reports the top-1
column for all drafters; the remaining columns are reported here without further
commentary. Calibration error in particular is reported plainly: it is a
diagnostic of the model's stated confidence, it was not part of the width
decision, and no claim in this paper rests on it.

The expert slice does not change which width leads. Across all four measures,
both pick policies, every set, and each format taken separately, the best-scoring
width under the expert slice is the same as under all drafters in 70 of 72 cells;
the two exceptions are log loss on BRO Traditional Draft, where the two smaller
widths exchange places. Expert-slice and per-format values accompany the artifact
data.

\begin{table}[tb]
\centering
\caption{Complete development results, all drafters, Premier and Traditional
Draft pooled. Top-1 and top-3 are percentages; the mean is the unweighted
three-set summary. Point estimates are shown. The 95\% percentile
cluster-bootstrap intervals resampling drafts ($B=1000$) have half-widths of at
most $\pm0.09$ percentage points on top-1, $\pm0.07$ on top-3, $\pm0.002$ nats on
log loss, and $\pm0.0008$ on calibration error; complete intervals accompany the
artifact data.}
\label{app:res-all}
{\small
\setlength{\tabcolsep}{8pt}
\begin{tabular}{llrrrrr}
\toprule
Set & Picks & $d$ & Top-1 & Top-3 & Log loss & ECE \\
\midrule
BRO & All picks & 128 & 50.63 & 83.66 & 1.3167 & 0.0515 \\
 &  & 256 & 50.83 & 83.58 & 1.3135 & 0.0546 \\
 &  & 512 & 49.89 & 83.35 & 1.3231 & 0.0459 \\
\addlinespace
 & Non-forced & 128 & 47.11 & 82.49 & 1.4104 & 0.0552 \\
 &  & 256 & 47.33 & 82.41 & 1.4071 & 0.0585 \\
 &  & 512 & 46.32 & 82.16 & 1.4173 & 0.0492 \\
\midrule
FDN & All picks & 128 & 60.15 & 90.80 & 1.0407 & 0.0464 \\
 &  & 256 & 60.37 & 91.13 & 1.0320 & 0.0444 \\
 &  & 512 & 60.56 & 91.10 & 1.0230 & 0.0275 \\
\addlinespace
 & Non-forced & 128 & 57.08 & 90.09 & 1.1209 & 0.0499 \\
 &  & 256 & 57.32 & 90.45 & 1.1115 & 0.0479 \\
 &  & 512 & 57.52 & 90.41 & 1.1018 & 0.0296 \\
\midrule
MSH & All picks & 128 & 56.71 & 88.04 & 1.1352 & 0.0536 \\
 &  & 256 & 56.67 & 88.22 & 1.1352 & 0.0586 \\
 &  & 512 & 56.79 & 88.15 & 1.1249 & 0.0435 \\
\addlinespace
 & Non-forced & 128 & 53.38 & 87.12 & 1.2225 & 0.0577 \\
 &  & 256 & 53.34 & 87.31 & 1.2225 & 0.0631 \\
 &  & 512 & 53.47 & 87.24 & 1.2114 & 0.0469 \\
\midrule
Mean & All picks & 128 & 55.83 & 87.50 & 1.1642 & 0.0505 \\
 &  & 256 & 55.96 & 87.65 & 1.1602 & 0.0525 \\
 &  & 512 & 55.75 & 87.53 & 1.1570 & 0.0390 \\
\addlinespace
 & Non-forced & 128 & 52.52 & 86.57 & 1.2513 & 0.0543 \\
 &  & 256 & 52.66 & 86.73 & 1.2470 & 0.0565 \\
 &  & 512 & 52.44 & 86.60 & 1.2435 & 0.0419 \\
\bottomrule
\end{tabular}
}
\end{table}

\subsection{Positional Behavior}
\label{app:positional}

\begin{figure}[tb]
\centering
\includegraphics[width=0.86\textwidth]{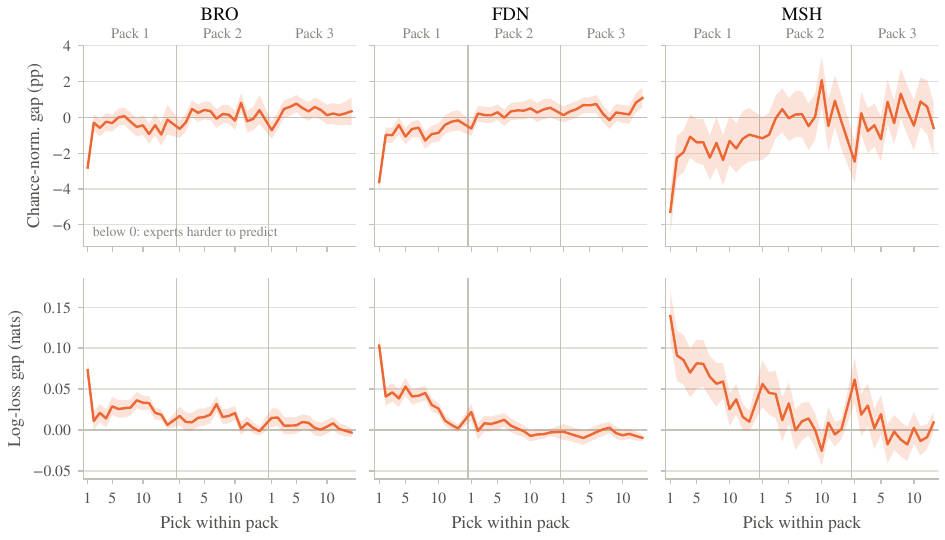}
\caption{Expert minus all-drafters difference across the draft, in
chance-normalized agreement (percentage points) and mean log loss (nats).
Experts are a win-rate bucket of at least 0.55 and a games-played bucket of at
least 100, per Section~\ref{sec:evaluation-measures}. Negative values mean expert
picks are harder to predict. The difference is largest at the first pick of the
draft and is no longer distinguishable from zero across most of pack 3. Bands are
95\% paired cluster-bootstrap intervals on the difference, resampling drafts
($B=1000$), with expert and all-drafter statistics recomputed within the same
resample.}
\label{fig:expert-gap}
\end{figure}

\begin{figure}[tb]
\centering
\includegraphics[width=0.86\textwidth]{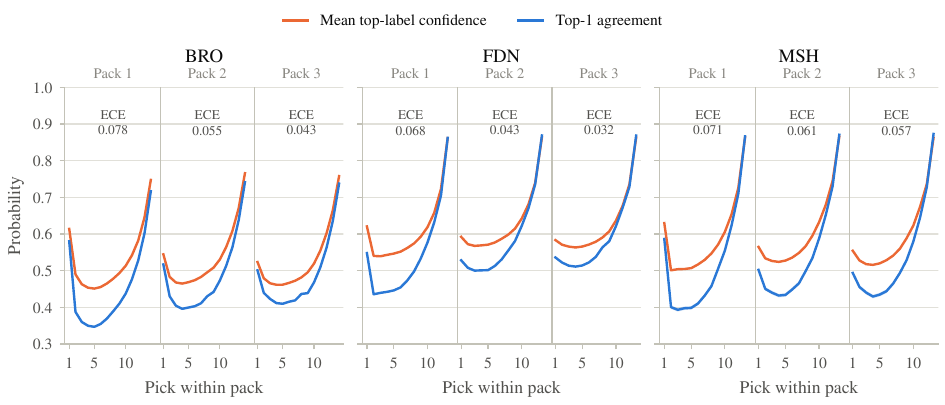}
\caption{Mean top-label confidence against top-1 agreement by draft position,
non-forced picks, DraftFM $d=256$, all drafters. The vertical gap is the model's
overconfidence. The annotated per-pack calibration error, 15 equal-mass bins,
coincides with that gap to within 0.0004, so the model is overconfident in every
bin rather than mixing over- and under-confidence. Overconfidence falls from pack
1 to pack 3 in all three sets. Bands are 95\% percentile cluster-bootstrap
intervals resampling drafts ($B=1000$).}
\label{fig:calibration-by-position}
\end{figure}

Two figures support sentences in
Section~\ref{sec:by-position-results}: Figure~\ref{fig:expert-gap} the expert
gap and Figure~\ref{fig:calibration-by-position} the calibration behavior. Each
caption states what its figure shows and the prose is not repeated here.

\clearpage

\renewcommand{\bibsection}{%
  \section*{References}%
  \drafttodo{Brian prose pass. The bibliography carries the cited literature.
  Footnotes that identify data, code, endpoints, or creator sources deliberately
  remain footnotes, and the artifact links live in
  Appendix~\ref{app:reproducibility}.}}

\bibliographystyle{plainnat}
\bibliography{references}

\end{document}